%% file: main.tex
\documentclass[10pt,journal,compsoc]{IEEEtran}
\IEEEoverridecommandlockouts
\usepackage{mydef}
\usepackage{cite}
\usepackage{amsmath,amssymb,amsfonts,bm}
\usepackage{algorithmic}
\usepackage{graphicx}
\usepackage{balance}
\usepackage{enumitem}
\usepackage{textcomp}
\usepackage{bbding} 
\usepackage{xcolor}
\usepackage{threeparttable}
\usepackage{algorithm}
\usepackage{multirow}
\usepackage{url}
\usepackage{booktabs}
\usepackage{array}
\usepackage{caption}
\usepackage{colortbl}
\usepackage{balance}
\usepackage{float,color}
\usepackage{romannum}

\def\BibTeX{{\rm B\kern-.05em{\sc i\kern-.025em b}\kern-.08em
    T\kern-.1667em\lower.7ex\hbox{E}\kern-.125emX}}
\begin{document}
\def\method{}
\newcommand{\R}[1]{{\color{black}{\sc} #1}}
\newcommand{\paratitle}[1]{\vspace{0.5ex}\noindent\textbf{\textit{#1}}}

\def \x {\mathbf{x}}
\def \X {{\mathbf{X}}}
\def \Y {\mathbf{Y}}
\def \y {\mathbf{y}}
\def \Z {\mathbf{Z}} \def \z {\mathbf{z}}

\def \ca {{\mathcal A}}
\def \cb {{\mathcal B}} \def \cd {{\mathcal D}} \def \cf {{\mathcal F}}
\def \cw {{\mathcal W}} \def \ck {{\mathcal K}} \def \cq {{\mathcal q}}
\def \cx {{\mathcal X}} \def \cm {{\mathcal M}} \def \cn {{\mathcal N}}
\def \cl {{\mathcal L}} \def \cR {{\mathcal R}} \def \cy {{\mathcal Y}}
\def \cu {{\mathcal U}} \def \ch {{\mathcal H}} \def \cp {{\mathcal P}}
\def \co {{\mathcal O}} \def \cs {{\mathcal S}} \def \ce {{\mathcal E}}
\def \cz {{\mathcal Z}} \def \cc {{\mathcal C}} \def \cg {{\mathcal G}}
\def \ct {{\mathcal T}} \def \ci {{\mathcal I}}
\def \proof {\noindent \emph{Proof.}\ \ }
\def \cv {{\mathcal V}}

%% indicate revision  in red with \rev{}
\newcommand{\revA}[1]{{\color{red} #1}}
\newcommand{\revB}[1]{{\color{blue} #1}}
\newcommand{\revC}[1]{{\color{green} #1}}
\newcommand{\revD}[1]{{\color{cyan} #1}}

\title{A Survey of Graph Neural Networks in Real world: Imbalance, Noise, Privacy and OOD Challenges}

\author{Wei Ju, \IEEEmembership{Member, IEEE}, Siyu Yi, \IEEEmembership{Member, IEEE}, Yifan Wang, Zhiping Xiao, Zhengyang Mao, \\Hourun Li, Yiyang Gu, Yifang Qin, Nan Yin, Senzhang Wang, \IEEEmembership{Member, IEEE}, \\Xinwang Liu, \IEEEmembership{Senior Member, IEEE,} Philip S. Yu, \IEEEmembership{Fellow, IEEE,} and Ming Zhang% <-this % stops a space
\IEEEcompsocitemizethanks{
\IEEEcompsocthanksitem Corresponding authors: Siyu Yi, Ming Zhang.
\IEEEcompsocthanksitem Wei Ju is with College of Computer Science, Sichuan University, Chengdu, China. (e-mail: juwei@scu.edu.cn)
\IEEEcompsocthanksitem Siyu Yi is with College of Mathematics, Sichuan University, Chengdu, China. (e-mail: siyuyi@scu.edu.cn)
\IEEEcompsocthanksitem Yifan Wang is with School of Information Technology $\&$ Management, University of International Business and Economics, Beijing, China.
(e-mail: yifanwang@uibe.edu.cn)
\IEEEcompsocthanksitem Zhiping Xiao is with Paul G. Allen School of Computer Science and Engineering, University of Washington, Seattle, WA, USA. (e-mail: patxiao@uw.edu)
\IEEEcompsocthanksitem Zhengyang Mao, Hourun Li, Yiyang Gu, Yifang Qin, and Ming Zhang are with School of Computer Science, Peking University, Beijing, China. (e-mail: zhengyang.mao@stu.pku.edu.cn, lihourun@stu.pku.edu.cn, yiyanggu@pku.edu.cn, qinyifang@pku.edu.cn, mzhang\_cs@pku.edu.cn)
\IEEEcompsocthanksitem Nan Yin is with Mohamed bin Zayed University of Artificial Intelligence, United Arab Emirates. (e-mail: yinnan8911@gmail.com)
\IEEEcompsocthanksitem Senzhang Wang is with the School of Computer Science and Technology, Central South University, Changsha, China. (e-mail: szwang@csu.edu.cn)
\IEEEcompsocthanksitem Xinwang Liu is with the College of Computer, National University of Defense Technology, Changsha, China. (e-mail: xinwangliu@nudt.edu.cn).
\IEEEcompsocthanksitem Philip S. Yu is with the Department of Computer Science, University of Illinois
at Chicago, Chicago, USA. (e-mail: psyu@uic.edu)
% \IEEEcompsocthanksitem Corresponding authors: Ming Zhang.
}% <-this % stops a space
}

\IEEEtitleabstractindextext{%

\input{1_abstract}

}

\maketitle
\input{2_introduction}
\input{3_taxonomy}
\input{4_preliminary}
\input{5_imbalance}

\input{6.1_label_noise}
\input{6.2_structure_noise}

\input{6.3_attribute_noise}

\input{7_privacy}
\input{8.1_ood_detection}
\input{8.2_ood_generalization}

\input{9_future_work}

% is not selected <OR> if conference mode is selected - because compsoc
% conference papers position the abstract like regular (non-compsoc)
\IEEEdisplaynontitleabstractindextext

% \fi
\IEEEpeerreviewmaketitle

\ifCLASSOPTIONcompsoc
  \section*{Acknowledgments}
\else
  \section*{Acknowledgment}
\fi

%The authors are grateful to the anonymous reviewers for critically reading this article and for giving important suggestions to improve this article. 

This paper is supported in part by National Natural Science Foundation of China under Grant 62306014 and 62276002, National Science Foundation under grants III-2106758 and POSE-2346158, Postdoctoral Fellowship Program (Grade A) of CPSF under Grant BX20250376 and BX20240239, China Postdoctoral Science Foundation under Grant 2024M762201, Sichuan Science and Technology Program under Grant 2025ZNSFSC1506 and 2025ZNSFSC0808, and Sichuan University Interdisciplinary Innovation Fund.

\ifCLASSOPTIONcaptionsoff
  \newpage
\fi



%\bibliographystyle{IEEEtran}
% \balance
\bibliographystyle{IEEEtran}
\bibliography{rec}


% \begin{IEEEbiography}
% [{\includegraphics[width=1in,height=1.25in]{bio/Ju.jpeg}}]
\smallskip\noindent\textbf{Wei Ju} is currently an associate professor with the College of Computer Science, Sichuan University. Prior to that, he worked as a postdoc and received his Ph.D. degree in the School of Computer Science from Peking University, in 2022. His current research interests focus on graph neural networks, bioinformatics, drug discovery, recommender systems, and spatio-temporal analysis. He has published more than 80 papers in top-tier venues.
% \end{IEEEbiography}

% \begin{IEEEbiography}
% [{\includegraphics[width=1in,height=1.25in]{bio/Yi.jpeg}}]
\smallskip\noindent\textbf{Siyu Yi} is currently a postdoc in Mathematics at Sichuan University. She received the Ph.D. degree in Statistics from Nankai University, in 2024. Her research interests focus on graph machine learning, statistical learning, and subsampling in big data. She has published more than 20 papers. 
% \end{IEEEbiography}

% \begin{IEEEbiography}
% [{\includegraphics[width=1in,height=1.25in]{bio/Yifan_Wang.jpg}}]
\smallskip\noindent\textbf{Yifan Wang} is currently an assistant professor in the School of Information Technology $\&$ Management, University of International Business and Economics. Prior to that, he received his Ph.D. degree in Computer Science from Peking University, in 2023. His research interests include graph neural networks, disentangled representation learning, drug discovery and recommender systems.
% \end{IEEEbiography}

% \begin{IEEEbiography}
% [{\includegraphics[width=1in,height=1.25in]{bio/Xiao.png}}]
\smallskip\noindent\textbf{Zhiping Xiao} received the Ph.D. degree from the Computer Science
Department, University of California at Los Angeles.
She is currently a postdoc with the
Paul G. Allen School of Computer Science and
Engineering, University of Washington. Her
current research interests include graph representation learning and social network analysis.
% \end{IEEEbiography}

% \begin{IEEEbiography}
% [{\includegraphics[width=1in,height=1.25in]{bio/Mao.jpg}}]
\smallskip\noindent\textbf{Zhengyang Mao} is currently a master's student at the School of Computer Science, Peking University. His research interests include graph representation learning, long-tailed learning and quantitative finance.
% \end{IEEEbiography}

% \begin{IEEEbiography}
% [{\includegraphics[width=1in,height=1.25in]{bio/Qin.JPG}}]
\smallskip\noindent\textbf{Yifang Qin} is an graduate student in School of Computer Science, Peking University, Beijing, China. Prior to that, he received the B.S. degree in school of EECS, Peking University. His research interests include graph representation learning and recommender systems.
% \end{IEEEbiography}

% \begin{IEEEbiography}
% [{\includegraphics[width=1in,height=1.25in]{bio/yinnan.jpg}}]
\smallskip\noindent\textbf{Nan Yin} is currently a postdoc at Mohamed bin Zayed University of Artificial Intelligence. He received the Ph.D. degree in School of Computer Science and Technology, National University of Defense Technology. His research interests includes transfer learning and graphs.
% \end{IEEEbiography}

% \begin{IEEEbiography}
% [{\includegraphics[width=1in,height=1.25in]{bio/Senzhang.png}}]
\smallskip\noindent\textbf{Senzhang Wang} received the Ph.D. degree from Beihang University. He is currently a Professor with School of Computer Science and Engineering, Central South University, Changsha. He has published over 100 papers on the top international journals and conferences. His current research interests include data mining and social network analysis.
% \end{IEEEbiography}

% \begin{IEEEbiography}
% [{\includegraphics[width=1.2in,height=1.25in,clip,keepaspectratio]{bio/xinwang Liu.png}}]
\smallskip\noindent\textbf{Xinwang Liu} received his Ph.D. degree from National University of Defense Technology (NUDT), in 2013. He is now Professor at School of Computer, NUDT. His current research interests include kernel learning, multi-view clustering and unsupervised feature learning. Dr. Liu has published 200+ peer-reviewed papers, including those in highly regarded journals and conferences such as TPAMI, TKDE, TIP, TNNLS, ICML, NeurIPS, CVPR, AAAI, IJCAI, etc. He is an Associate Editor of IEEE T-NNLS and Information Fusion Journal.
% \end{IEEEbiography}

% \begin{IEEEbiography}
% [{\includegraphics[width=1in,height=1.25in]{bio/Luo.jpg}}]
% {Xiao Luo} is a postdoctoral researcher in Department of Computer Science, University of California, Los Angeles, USA. Prior to that, he received the Ph.D. degree in School of Mathematical Sciences from Peking University, Beijing, China and the B.S. degree in Mathematics from Nanjing University, Nanjing, China, in 2017. 
% His research interests includes machine learning on graphs, image retrieval, statistical models and bioinformatics. 
% \end{IEEEbiography}

% \begin{IEEEbiography}
% [{\includegraphics[width=1in,height=1.25in]{bio/Yu.png}}]
\smallskip\noindent\textbf{Philip S. Yu} received the Ph.D. degree in E.E. from Stanford University. He is a distinguished professor in Computer Science at the University of Illinois at Chicago. He a recipient of ACM SIGKDD 2016 Innovation Award for his influential research and scientific contributions on mining, fusion and anonymization of big data, the IEEE Computer Societys 2013 Technical Achievement Award for pioneering and fundamentally innovative contributions to the scalable indexing, querying, searching, mining and anonymization of big data. He was Editor-in-Chiefs of ACM Transactions on Knowledge Discovery from Data (2011-2017) and IEEE Transactions on Knowledge and Data Engineering (2001-2004).
% \end{IEEEbiography}

% \begin{IEEEbiography}
% [{\includegraphics[width=1in,height=1.25in]{bio/Zhang.jpeg}}]
\smallskip\noindent\textbf{Ming Zhang} received her Ph.D. degree in Computer Science from Peking University. She is a full professor at the School of Computer Science, Peking University. Prof. Zhang is a member of Advisory Committee of Ministry of Education in China and the Chair of ACM SIGCSE China. She is one of the fifteen members of ACM/IEEE CC2020 Steering Committee. She has published more than 200 research papers on Text Mining and Machine Learning in the top journals and conferences. She won the best paper of ICML 2014 and best paper nominee of WWW 2016 and ICDM 2022. 
% Prof. Zhang is the leading author of several textbooks on Data Structures and Algorithms in Chinese, and the corresponding course is awarded as the National Elaborate Course, National Boutique Resource Sharing Course, National Fine-designed Online Course, National First-Class Undergraduate Course by MOE China.
% \end{IEEEbiography}

\end{document}

%% file: 1_abstract.tex
\begin{abstract}
Graph-structured data exhibits universality and widespread applicability across diverse domains, such as social network analysis, biochemistry, financial fraud detection, and network security. Significant strides have been made in leveraging Graph Neural Networks (GNNs) to achieve remarkable success in these areas. However, in real-world scenarios, the training environment for models is often far from ideal, leading to substantial performance degradation of GNN models due to various unfavorable factors, including imbalance in data distribution, the presence of noise in erroneous data, privacy protection of sensitive information, and generalization capability for out-of-distribution (OOD) scenarios. To tackle these issues, substantial efforts have been devoted to improving the performance of GNN models in practical real-world scenarios, as well as enhancing their reliability and robustness. In this paper, we present a comprehensive survey that systematically reviews existing GNN models, focusing on solutions to the four mentioned real-world challenges including imbalance, noise, privacy, and OOD in practical scenarios that many existing reviews have not considered. Specifically, we first highlight the four key challenges faced by existing GNNs, paving the way for our exploration of real-world GNN models. Subsequently, we provide detailed discussions on these four aspects, dissecting how these solutions contribute to enhancing the reliability and robustness of GNN models. Last but not least, we outline promising directions and offer future perspectives in the field.
\end{abstract}

%% file: 2_introduction.tex
\section{Introduction}

\IEEEPARstart{G}{raph-structured} data, characterized by nodes and edges that represent interconnected entities and relationships, possesses inherent complexity and versatility. The interconnected nature of graphs allows them to model a wide range of real-world scenarios where entities and their interactions play a crucial role. Analyzing graph data is of paramount importance as it enables us to gain insights into intricate patterns, uncover hidden structures, and understand the dynamics of interconnected systems~\cite{wu2020comprehensive,ju2024comprehensive}. The applicability of graph data extends across various domains; for instance, in social network analysis, graphs can represent relationships between individuals~\cite{xu2024temporal}, in bioinformatics, molecular structures can be modeled as graphs~\cite{yao2024property}, and transportation networks can be also expressed as graphs to optimize routes and logistics~\cite{li2024survey}. 
% These examples underscore the significance and broad utility of graph data analysis in addressing diverse challenges across different fields.

Recently, the landscape of graph data analysis has been significantly shaped by the widespread adoption and remarkable success of Graph Neural Networks (GNNs)~\cite{kipf2017semi,velivckovic2018graph,hamilton2017inductive,ju2024survey}. GNNs have emerged as a cornerstone in graph learning, demonstrating exceptional performance in various applications. The fundamental idea behind GNNs lies in their ability to capture complex relationships within graph-structured data by iteratively aggregating and updating information from neighboring nodes~\cite{gilmer2017neural}. This enables GNNs to learn meaningful representations of nodes, capturing both local and global patterns within the graph~\cite{wu2020comprehensive}. The versatility and effectiveness of GNNs are prominently demonstrated in various real-world applications. In e-commerce, platforms like Alibaba~\cite{zhu2019aligraph} leverage GNNs to comprehend user behavior, thereby enabling personalized product recommendations and enhancing overall user engagement. Social media such as Pinterest~\cite{pal2020pinnersage} utilize GNNs for content recommendation, successfully connecting users with relevant and appealing content. Additionally, GNNs achieve remarkable success in scenarios such as simulating complex physical systems~\cite{yuan2024egode,luo2023hope} and accelerating drug discovery processes~\cite{xu2023graph,ju2023few}.

\begin{figure}
\centering
\includegraphics[width=\linewidth]{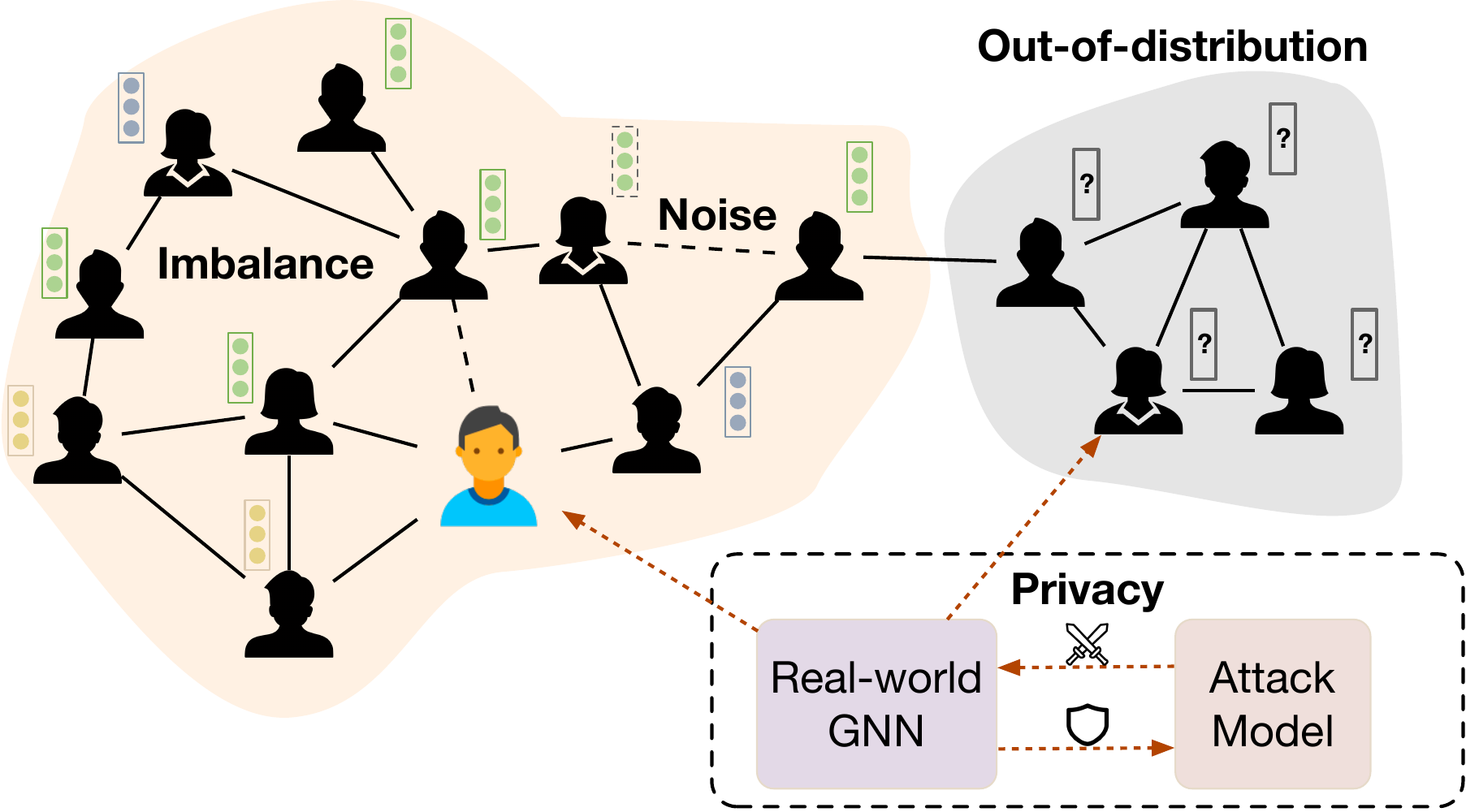}
\caption{An illustrative example of GNN models handling practical social network scenarios. User data extracted from real-world platforms typically exhibit long-tailed distributions, indicating there are widespread mainstream user types alongside lots of rare genres. The interactions among users may be influenced by structural noises and the presence of fake labels. Moreover, the practical GNN models are confronted with attack models and user information leakage issues. The generalization of models from existing business scenarios to novel environments also introduces OOD concerns. 
% To tackle these real-world challenges, innovative model structures and practical techniques are required to enhance the robustness and reliability of GNN models.
}
% imbalances among node categories, structural noise, and practical privacy concerns. Moreover, the model generalization from existing domains to novel environments also causes OOD issues. To tackle these real-world challenges, innovative model structures and practical techniques are required to enhance the robustness and reliability of GNN models.}
\label{fig:example}
\end{figure}

Despite the outstanding performance exhibited by current GNN models, it is crucial to acknowledge that their training typically occurs within an idealized environment, where the training data is clean, standardized, and comprehensive. However, in real-world scenarios, GNN models typically face various challenges that significantly compromise their performance and may even lead to model collapse~\cite{zhang2024trustworthy,dai2024comprehensive}. This discrepancy between idealized training conditions and real-world challenges poses a critical issue in the deployment of GNNs. For example, in the fraud detection of financial transactions~\cite{choi2025unveiling}, compared with non-fraud cases, the scarcity of fraud cases leads to an imbalance dataset. Due to this \textbf{imbalance} problem, GNNs may have difficulty effectively learning patterns related to fraud. In bioinformatics~\cite{li2025graph}, experimental errors or anomalies may introduce \textbf{noise}, making it difficult for GNNs to accurately predict molecular structures or identify potential patterns in biological data. In social network analysis~\cite{lei2025achieving}, GNNs models needs to strike a balance between extracting valuable information from the network and preserving user \textbf{privacy}. Furthermore, in network security~\cite{lin2025conformal}, GNNs used to detect network threats may encounter struggle when facing novel and previously unseen \textbf{out-of-distribution (OOD)} attacks. The illustrative example in Figure~\ref{fig:example} further elucidates the challenges encountered in real-world social network scenarios. 
% These real-world examples underscore the vulnerability of GNN models to various unfavorable challenges, highlighting the importance of developing reliable and robust solutions for GNNs.
%  in practical applications.

To cope with the myriad challenges that GNN models face in real-world scenarios, researchers have dedicated significant efforts to address these adverse factors. To comprehensively and systematically summarize the methodologies employed in real-world scenarios, we present a thorough survey in this paper. This survey primarily focuses on the solutions devised for GNN models confronting four prevalent real-world conditions: \emph{Imbalance, Noise, Privacy, and Out-of-Distribution}. By consolidating existing research endeavors, this survey aims to provide a comprehensive overview of the current landscape. Additionally, we aim to present prospective research frontiers that can guide researchers in reviewing, summarizing, and formulating future strategies to enhance the reliability and robustness of GNN models in practical applications.

\smallskip
\noindent\textbf{Differences between this survey and existing ones}.
Until now, there have been several other literature reviews that have delved into real-world GNNs from different aspects~\cite{dai2024comprehensive,zhang2024trustworthy,wu2022trustworthy,oneto2022towards,li2022recent}, and they are closely relevant to our research. While these surveys share relevance with our work, they also exhibit distinctions in their specific focuses. For example, Wu et al.~\cite{wu2022trustworthy} focus on three aspects of GNN models: reliability, explainability, and privacy. Dai et al.~\cite{dai2024comprehensive} conduct a more detailed discussion covering privacy, robustness, fairness, and explainability. Zhang et al.~\cite{zhang2024trustworthy}, building upon the foundation laid by \cite{dai2024comprehensive}, explore the emerging topics of accountability and environmental well-being. These three concurrent works center their themes around trustworthy GNNs, approaching from the perspective of creating more reliable AI systems. Different from theses works, our survey originates from real-world considerations, concentrating on practical scenarios. Further, Oneto et al.~\cite{oneto2022towards}, expanding on the trustworthy foundation, encompass more macroscopic elements such as automated operations with guarantees on graphs, aiming for more intelligent and responsible GNN models. To the best of our knowledge, the survey most closely related to ours is \cite{li2022recent}, which summarizes reliable graph learning from the aspects of inherent noise, distribution shift, and adversarial attack. Besides these, our survey also address the prevalent issues of data imbalance and privacy in real-world scenarios. It is worth noting that their survey~\cite{li2022recent} only covers methods up to 2022, lacking coverage of the latest developments in the past three years.

\smallskip
\noindent\textbf{Our Contribution.} 
This survey aims to comprehensively summarize the advancements of GNN models in the real world while also paving the way for future exploration. It serves as a valuable resource for both researchers and practitioners by providing them with an overview and the latest developments in GNNs in practical scenarios. The key contributions of this survey are highlighted below:

\begin{itemize}[leftmargin=*]
    \item \textbf{Systematic Taxonomy.} A novel taxonomy is proposed to systematically categorize existing real-world GNN models, focusing primarily on the models to address imbalance, noise, privacy, and out-of-distribution issues, and presenting representative methods.
    \item \textbf{Extensive Review.} For each category covered in this survey, we summarize its basic principles and components, and provide detailed insights into representative algorithms, followed by a systematic discussion of their findings.
    \item \textbf{Future Perspectives.} 
    We identify limitations and confront challenges associated with current real-world GNN models, and outline potential research directions, offering a novel perspective on future avenues of study.
\end{itemize}

\tikzstyle{leaf}=[draw=black, %边框
    rounded corners,minimum height=1.2em,
    text width=30.08em, edge=black!10, 
    %fill=hidden-orange!40,
    text opacity=1, align=center,
    fill opacity=.3,  text=black,font=\scriptsize,
    inner xsep=3pt, inner ysep=1pt,
    ]
\tikzstyle{leaf1}=[draw=black, %边框
    rounded corners,minimum height=1.2em,
    text width=6.5em, edge=black!10, 
    text opacity=1, align=center,
    fill opacity=.5,  text=black,font=\scriptsize,
    inner xsep=3pt, inner ysep=1pt,
    ]
\tikzstyle{leaf2}=[draw=black, %边框
    rounded corners,minimum height=1.2em,
    text width=5.5em, edge=black!10, 
    text opacity=1, align=center,
    fill opacity=.8,  text=black,font=\scriptsize,
    inner xsep=3pt, inner ysep=1pt,
    ]
\tikzstyle{leaf3}=[draw=black, %边框
    rounded corners,minimum height=1.2em,
    text width=6.28em, edge=black!10, 
    text opacity=1, align=center,
    fill opacity=.8,  text=black,font=\scriptsize,
    inner xsep=3pt, inner ysep=1pt,
    ]
\begin{figure*}[ht]
\centering
\begin{forest}
  for tree={
  forked edges,
  grow=east,
  reversed=true,
  anchor=base west,
  parent anchor=east,
  child anchor=west,
  base=middle,
  font=\scriptsize,
  rectangle,
  draw=black, %hiddendraw 所有边框
  edge=black!50, 
  rounded corners,
  align=center,
  minimum width=2em,
  s sep=5pt,
  inner xsep=3pt,
  inner ysep=1pt
  },
  % where level=1{align=center}{},
  % where level=2{text width=6em,font=\scriptsize}{},
  % where level=3{font=\scriptsize}{},
  % where level=4{font=\scriptsize}{},
  % where level=5{font=\scriptsize}{},
  [Graph Learning in the Real World,rotate=90,anchor=north,edge=black!50,fill=myblue,draw=black
    %%%
    [Imbalance,edge=black!50, fill=mypurple, minimum height=1.2em
    	[Re-balancing, leaf1,text width=7.8em,fill=mypurple
	    [GraphSMOTE \cite{zhao2021graphsmote}{,} ImGAGN \cite{qu2021imgagn}{,} GraphENS \cite{park2022graphens}{,} SNS \cite{gao2023semantic}{,} C$^3$GNN \cite{ju2025cluster}{,} ReNode \cite{chen2021topology}{,} TAM \cite{song2022tam} ,leaf,text width=34.68em, fill=mypurple]
            ]
	[Augmentation-based,leaf1,text width=7.8em,fill=mypurple
	    [GraphMixup \cite{wu2022graphmixup}{,} G$^2$GNN \cite{wang2022imbalanced}{,} CM-GCL \cite{qian2022co}{,}\\
        SOLTGNN \cite{liu2022size}{,} RAHNet \cite{mao2023rahnet}{,}  GNN-INCM \cite{huang2022graph}{,}LTE4G \cite{yun2022lte4g} ,leaf,text width=34.68em, fill=mypurple]
	]
	[Module Improvement, leaf1,text width=7.8em,fill=mypurple
	    [ImGCL \cite{zeng2023imgcl}{,} INS-GNN \cite{juan2023ins}{,} GNN-CL \cite{li2024graph}{,} RAHNet~\cite{mao2023rahnet}{,} GraphDIVE~\cite{hu2022graphdive}{,} CoMe~\cite{yi2023towards}, leaf,text width=34.68em, fill=mypurple]
	]
    ]
    %%%
    [Noise,edge=black!50, fill=myred, minimum height=1.2em,
        [Label Noise, leaf2, fill=myred
            [Loss Correction, leaf1,fill=myred
                 [NRGNN \cite{dai2021nrgnn}{,} DND-Net \cite{ding2024divide}{,} PIGNN \cite{du2023noise}{,} CP \cite{zhang2020adversarial}{,} RTGNN \cite{qian2023robust}{,} GraphCleaner \cite{li2023graphcleaner}, leaf,fill=myred]
            ]
            [Label Correction, leaf1,fill=myred
                 [UnionNET \cite{li2021unified}{,} GNN Cleaner \cite{xia2023gnn}{,} ERASE \cite{chen2024erase}{,} LP4GLN \cite{cheng2023label}{,} CGNN \cite{yuan2023learning}, leaf,fill=myred]
            ]
        ]
        [Structure Noise, leaf2, fill=myred
            [Metric Learning, leaf1, fill=myred
                 [GRCN \cite{yu2021graph}{,} GNNGuard \cite{zhang2020gnnguard}{,} GDC \cite{gasteiger2019diffusion}{,} GLCN \cite{jiang2019semi}{,} IDGL \cite{chen2020iterative}{,} SLAPS \cite{fatemi2021slaps}, leaf, edge=black!50, fill=myred]
            ] 
            [Sampling-based, leaf1, fill=myred
                 [DropEdge \cite{rong2020dropedge}{,} DropCONN \cite{chen2020enhancing}{,} PTDNet \cite{luo2021learning}{,} FastGCN \cite{chen2018fastgcn}{,} NeuralSparse \cite{zheng2020robust}, leaf,fill=myred]
            ] 
            [Direct Optimization, leaf1, fill=myred
                 [TO-GCN \cite{yang2019topology}{,} 
                 Pro-GNN \cite{jin2020graph}{,} 
                 Gosch et al.~\cite{gosch2023adversarial}{,}
                 PTA \cite{dong2021equivalence}{,} 
                 RLP \cite{he2022structural}{,} 
                 PAMT \cite{chen2024pamt}, leaf,fill=myred]
            ]  
        ]
        [Attribute Noise, leaf2, fill=myred
            [Adversarial Attack  \\ and Defense, leaf1,fill=myred
                 [Nettack \cite{zugner2018adversarial}{,} GraphAT~\cite{feng2019graph}{,} GCNVAT~\cite{sun2019virtual}{,} BVAT~\cite{deng2023batch}{,} GCORN~\cite{abbahaddou2024bounding}, leaf, edge=black!50, fill=myred]
            ]
            [Loss Refinement, leaf1,fill=myred
                 [T2-GNN~\cite{huo2023t2}{,}  MQE \cite{li2024noise}{,} BRGCL \cite{wang2022bayesian}, leaf, edge=black!50, fill=myred]
            ]
        ]
    ]
    %%%
    [Privacy,edge=black!50, fill=mygreen, minimum height=1.2em,
    	[Privacy Attack, leaf1,text width=7em,fill=mygreen
	    [MIAGraph~\cite{olatunji2021membership}{,} He et al.~\cite{he2021node}{,} Duddu et al.~\cite{duddu2020quantifying}{,} GraphMI~\cite{zhang2021graphmi}{,} 
        Wu et al.~\cite{wu2022model}{,} 
        Shen et al.~\cite{shen2022model}, leaf,text width=36.35em, fill=mygreen]
            ]
	[Privacy Preservation, leaf1,text width=7em,fill=mygreen  [
        DPNE~\cite{xu2018dpne}{,} PrivGnn~\cite{olatunji2021releasing}{,} DP-GNN~\cite{mueller2022differentially}{,} KProp~\cite{sajadmanesh2021locally}{,} GERAI~\cite{zhang2021graph}{,} GAL~\cite{liao2021information}{,} APGE~\cite{li2020adversarial}{,} \\
        DP-GCN~\cite{hu2022learning}{,} SpreadGNN~\cite{he2021spreadgnn}{,} D-FedGNN~\cite{pei2021decentralized}{,} GraphErase~\cite{chen2022graph} MIAGraph~\cite{olatunji2021membership},
        leaf,text width=36.35em, fill=mygreen]
	]
    ]
    %%%
    [OOD,edge=black!50, fill=myyellow, minimum height=1.2em,
    	[OOD Detection,leaf2, fill=myyellow, text width=7em, 
            [Propagation-based, leaf1, fill=myyellow, text width=7em, 
                 [GPN~\cite{stadler2021graph}{,} GNNSage~\cite{wu2023energy}{,} OODGAT~\cite{song2022learning}{,} OSSNC~\cite{huang2022end},leaf, fill=myyellow, text width=28.25em,]
            ]  
            [Classification-based, leaf1, fill=myyellow, text width=7em, 
                 [AAGOD~\cite{guo2023data}{,} BWGNN~\cite{tang2022rethinking}{,} GKDE~\cite{zhao2020uncertainty}{,} iGAD~\cite{zhang2022dual}, leaf,fill=myyellow, text width=28.25em,]
            ]  
            [Self-supervised \\ Learning-based, leaf1, fill=myyellow, text width=7em, 
                 [GLocalKD~\cite{ma2022deep}{,} GOOD-D~\cite{liu2023good}{,} GRADATE~\cite{duan2023graph}{,} GLADC~\cite{luo2022deep}{,} \\ GraphDE~\cite{li2022graphde}{,}
                 OCGIN~\cite{zhao2023using}{,} OCGTL~\cite{qiu2023raising}{,} GOODAT~\cite{wang2024goodat}, leaf,fill=myyellow, text width=28.25em]
            ]  
         ]
         [OOD Generalization, leaf2, fill=myyellow, text width=7em, 
            [Subgraph-based, leaf1, fill=myyellow, text width=7em, 
            	[CAL~\cite{sui2022causal}{,} CIGA~\cite{chen2022learning}{,} StableGNN~\cite{fan2023generalizing}{,} SizeShiftReg~\cite{buffelli2022sizeshiftreg}{,} \\ GIL~\cite{li2022learning}{,} MoleOOD~\cite{yang2022learning}{,} LiSA~\cite{yu2023mind}{,} EERM~\cite{wu2022handling},leaf,text width=28.25em,fill=myyellow]
            ] 
            [Adversarial Learning, leaf1, fill=myyellow, text width=7em, 
                 [GraphAT~\cite{feng2019graph}{,} CAP~\cite{xue2021cap}{,} AIA~\cite{sui2023unleashing}{,} LECI~\cite{gui2023joint}{,} WT-AWP~\cite{wu2023adversarial}{,} DEAL~\cite{yin2022deal}, leaf,text width=28.25em,fill=myyellow]
            ]
         ]
    ]
  ]
\end{forest}
\caption{An overview of the taxonomy for existing GNN models in real world.}
\label{fig:taxonomy_of_DEGL}
\end{figure*}
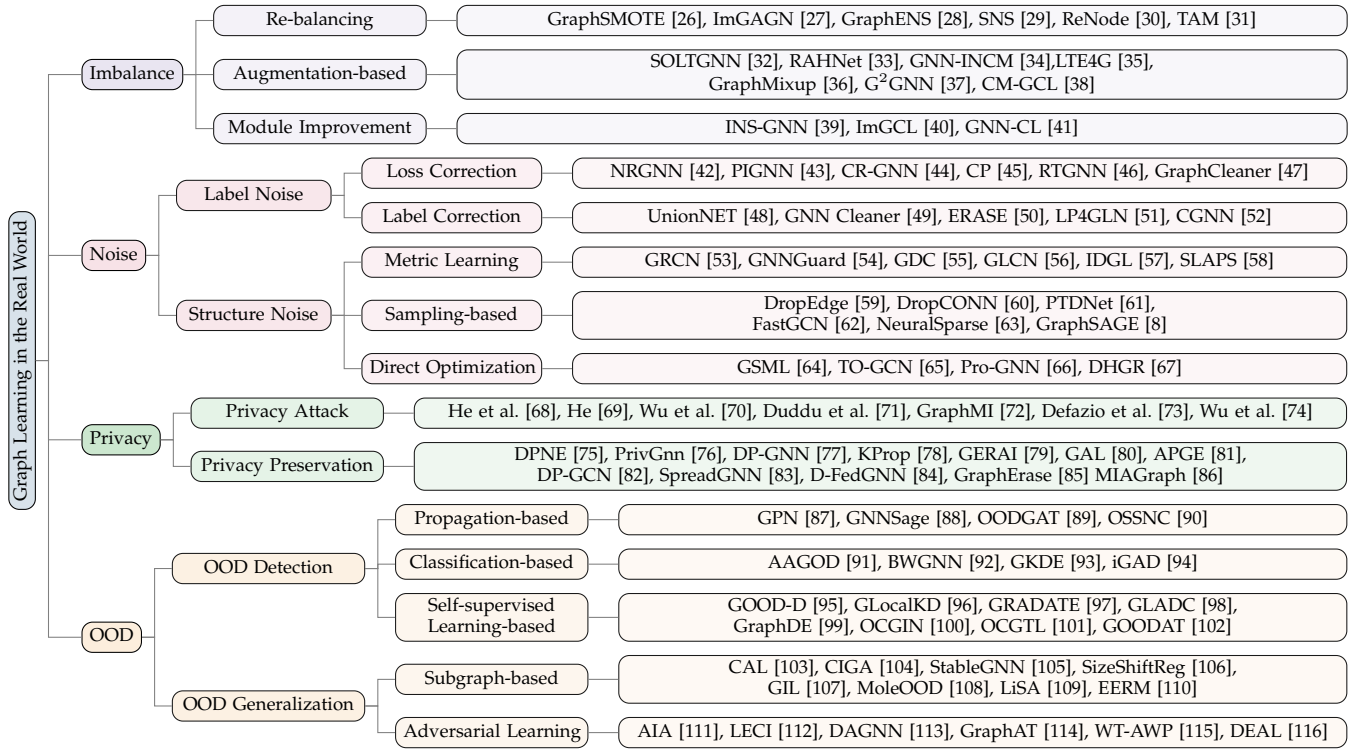

%% file: 3_taxonomy.tex
\section{Taxonomy}

To gain deeper insights of GNN models in real-world scenarios, we have spotlighted key research efforts, delved into their motivations, and concisely summarized their primary technical contributions. The overall structure of the paper is depicted in Figure \ref{fig:taxonomy_of_DEGL}. This survey establishes a novel taxonomy, categorizing these works into four distinct classes: \emph{Imbalance, Noise, Privacy, and Out-Of-Distribution}. These categories serve as a comprehensive framework for reviewing and analyzing these works across diverse scenarios. We will provide a brief overview of these four real-world factors:

\begin{itemize}[leftmargin=*]
% \item \paratitle{Imbalance} in graph data refers to a situation where the distribution of classes or labels within a graph is highly uneven~\cite{zhao2021graphsmote}. 
% The key idea behind addressing imbalance is to ensure that the learning process is not skewed toward the majority class, enabling the model to effectively capture patterns associated with the minority class. This involves devising strategies to prevent the dominant nodes or classes from overshadowing the contributions of the minority nodes or classes during the training process. To tackle the class imbalance issue, three fundamental principles are generally employed. The first is \emph{Re-balancing} strategy, aiming to achieve a balanced situation for different categories of samples or loss functions through specific techniques~\cite{zhao2021graphsmote,chen2021topology}. The second is \emph{Augmentation-based} strategy, designed to enhance model training with additional information~\cite{yun2022lte4g,wang2022imbalanced}. The last is \emph{Module Enhancement} strategy, which aims to enhance the representation learning of network modules in imbalanced learning~\cite{zeng2023imgcl,hu2022graphdive}.

\item \paratitle{Imbalance} in graph data refers to the uneven distribution of class labels in a graph~\cite{zhao2021graphsmote}. Solving this problem is crucial for preventing the learning process from being biased towards the majority class, which will hinder the model's ability to capture minority class patterns. To alleviate this situation, three main strategies are usually employed. The first is the re-balancing strategy, which aims to reduce class dominance~\cite{zhao2021graphsmote,chen2021topology} by adjusting sample distributions or loss functions. The second strategy is augmentation-based, which aims to support model training by introducing extra data or structural information~\cite{yun2022lte4g,wang2022imbalanced}. The last strategy is module enhancement~\cite{zeng2023imgcl,hu2022graphdive} to increase the model's ability to learn discriminative features under imbalance.

% \item \paratitle{Noise} in graph data refers to the presence of irrelevant, incorrect, or misleading information within the graph, which can adversely impact the performance of GNN models~\cite{dai2021nrgnn}. The fundamental idea behind addressing noise is to develop strategies that mitigate the impact of erroneous information during the training process, accurately capturing underlying patterns within the graph. This involves distinguishing between two primary types of noise: \emph{Label Noise} and \emph{Structure Noise}. Label noise~\cite{dai2021nrgnn,xia2023gnn} pertains to inaccuracies or errors in the assigned labels of nodes or edges within the graph. This can occur due to human error during data labeling or inconsistencies in the data collection process. Structure noise~\cite{jiang2019semi,rong2020dropedge,yang2019topology}, on the other hand, relates to inconsistencies or inaccuracies in the topology of the graph. This could include missing or erroneous connections between nodes, disrupting the genuine relationships within the graph.

\item \paratitle{Noise} in graph data denotes incorrect, irrelevant, or misleading information that can impair GNN performance~\cite{dai2021nrgnn}, generally classified into three types: \emph{Label Noise}, \emph{Structure Noise}, and \emph{Attribute Noise}. 
% Some techniques have been developed to reduce the influence of such noise and better capture the graph’s underlying patterns. 
The first is caused by label allocation errors due to annotation errors or inconsistent data, which is typically addressed through {loss correction} and {label correction}~\cite{dai2021nrgnn,xia2023gnn}. 
The second arises from inaccurate or missing edges and distorted graph topology, which is commonly mitigated through metric learning, {sampling-based methods} and {direct optimization}~\cite{jiang2019semi,rong2020dropedge,yang2019topology}. 
The third arises from manual entry errors or intentional manipulation, typically addressed by ensuring robust representation learning through adversarial training~\cite{zugner2018adversarial} and loss refinement~\cite{li2024noise}.

% \item \paratitle{Privacy} in graph data concerns safeguarding sensitive information associated with nodes or edges within a graph, ensuring that the confidentiality and integrity of such data are preserved~\cite{zhang2024survey}. For effective graph learning, privacy becomes a critical consideration due to the potential exposure of personal or confidential details during the training and inference phases. The fundamental idea behind addressing privacy concerns is to develop strategies that strike a balance between meaningful insights extracted from the graph and the sensitive information protection. This involves recognizing two primary categories: \emph{Privacy Attack} and \emph{Privacy Preservation}. Privacy attacks~\cite{olatunji2021membership,duddu2020quantifying,wu2022model} involve attempts to exploit vulnerabilities in the graph data to uncover sensitive information about individuals or entities. 
% % Defending against privacy attacks requires implementing techniques to obfuscate or protect sensitive information, making it more challenging for adversaries to infer private details. 
% Privacy preservation~\cite{hu2022learning,li2020adversarial,mcmahan2017communication} focuses on developing mechanisms and techniques to safeguard sensitive information within the graph, ensuring that even with access to certain portions of the data, it remains challenging to disclose private details.

\item \paratitle{Privacy} in graph data involves protecting sensitive information tied to nodes or edges to ensure confidentiality during training and inference processes~\cite{zhang2024survey}. Since graph learning usually deals with personal or confidential data, protecting privacy is of crucial importance. The solution aims to balance practicality and protection, and is usually divided into \emph{Privacy Attack} and \emph{Privacy Preservation}. Privacy attack uses the model or data privacy loophole to reveal hidden information~\cite {olatunji2021membership,duddu2020quantifying, wu2022model}. And privacy preservation is focused on by adversarial learning, the federated training, and other technical defense~\cite{hu2022learning,li2020adversarial,mcmahan2017communication}.

% \item \paratitle{Out-Of-Distribution (OOD)} in graph data refers to instances or patterns that deviate significantly from the data distribution encountered during the model's training phase~\cite{li2022ood}. In graph learning, OOD scenarios involve encountering graph instances that differ substantially from those observed during the model training. The fundamental idea behind addressing OOD in graph data is to equip models with the capability to identify and handle instances that fall outside the distribution seen during training. This involves recognizing two primary categories: \emph{OOD Detection} and \emph{OOD Generalization}. OOD detection~\cite{stadler2021graph,li2022graphde,liu2023good} focuses on developing techniques to identify instances in the graph data that do not conform to the learned distribution during training. This often involves leveraging anomaly detection methods or incorporating uncertainty estimation mechanisms to flag instances that exhibit characteristics inconsistent with the training data. OOD generalization~\cite{sui2022causal,wu2019domain} aims to enhance the model's ability to make accurate predictions on instances that deviate from the training distribution. The goal is to enable the GNN model to make reliable predictions even in the face of novel, unseen graph instances.

\item \paratitle{Out-of-distribution (OOD)} in graph data are instances that differ significantly from the distribution seen during training~\cite{li2022ood}. In graph learning, OOD scenarios require the model to recognize and adapt to unseen graph instances. Solving this problem involves two key tasks: \emph{OOD Detection} and \emph{OOD Generalization}. OOD detection~\cite {stadler2021graph,li2022graphde,liu2023good} focuses on the recognition beyond the distribution of the training data points, usually use anomaly detection or uncertainty estimation techniques. Instead, OOD generalization~\cite {sui2022causal,wu2019domain} aims to improve the robustness of the models, to new, unseen graph for accurate prediction.
\end{itemize}

%% file: 4_preliminary.tex
\section{Preliminary}
% To facilitate the introduction of real-world GNN models in the following sections, 
% In this section, we first provide definitions of graphs, the fundamental principles of GNNs, and an overview of graph computational tasks.

\subsection{Graph}
Given a graph denoted as $\mathcal{G} = (\mathcal{V}, \mathcal{E})$, where $\mathcal{V}=\{v_1,\ldots,v_{|\mathcal{V}|}\}$ is the node set and $\mathcal{E}$ is the edge set which can be represented by an adjacency matrix $\mathbf{A} \in \mathbb{R}^{|\mathcal{V}| \times |\mathcal{V}|}$, where $\mathbf{A}_{ij}=1$ if $(v_i, v_j)\in\mathcal{E}$, otherwise $\mathbf{A}_{ij}=0$. Each node $v_i$ in the graph is associated with a feature vector $\mathbf{x}_i \in \mathbb{R}^d$, constituting the feature matrix of graph $\mathbf{X} \in \mathbb{R}^{|\mathcal{V}| \times d}$, where $d$ presents the number of dimensions in the features. Therefore, we can also represent a graph as $\mathcal{G}=\{\mathbf{X}, \mathbf{A}\}$, in which $\mathbf{Y}$ denotes the label vector for the labeled nodes or graphs.

\subsection{Graph Neural Networks}
Graph Neural Networks (GNNs)~\cite{kipf2017semi,velivckovic2018graph,hamilton2017inductive} represent a class of neural network architectures specifically tailored for learning representations of the graph's components—its nodes, edges and even entire graph—to capture the complex relationships and structures within the graph. A central mechanism of GNNs is the message-passing paradigm~\cite{gilmer2017neural}, where the embedding of a node $v_i$ is iteratively updated through  interactions with its neighbors, denoted as:
\begin{equation}
\begin{split}
    \mathbf{h}_i^{(l)} &= \text{GNN}^{(l)}\Bigl(\mathbf{h}_i^{(l-1)}, \bigl\{\mathbf{h}_j^{(l-1)}\bigr\}_{v_j \in \mathcal{N}(v_i)}\Bigr)\\
    & = \mathcal{C}^{(l)}\Bigl(\mathbf{h}_i^{(l-1)}, \mathcal{A}^{(l)}\Bigl(\bigl\{\mathbf{h}_j^{(l-1)}\bigr\}_{v_j \in \mathcal{N}(v_i)}\Bigr)\Bigr),
\end{split}
\end{equation}
where $\mathbf{h}_i^{(l)}$ indicates the embedding of $v_i$ at layer $l\in\{1,\dots,L\}$ and $\mathcal{N}(v_i)$ denotes the neighbors of $v_i$ derived from $\mathbf{A}$. $\mathcal{A}^{(l)}$ and $\mathcal{C}^{(l)}$ are the message aggregating and embedding updating functions at layer $l$, respectively. Finally, the node-level representation is $\mathbf{h}_i = \mathbf{h}_i^{(L)}$ at layer $L$, while the graph-level representation can be attained by a READOUT aggregation function, defined as: 
% aggregating all node representations at layer $L$ with a readout function. Formally it can be defined as:
\begin{equation}
    \mathbf{h}_{\mathcal{G}} = \text{READOUT}(\{\mathbf{h}_i^{(L)}\}_{v_i\in\mathcal{V}}),
\end{equation}
% where $h_{\mathcal{G}}$ is the graph-level representation, 
where $\text{READOUT}$ could be averaging or other graph-level pooling functions depending on the model~\cite{lee2019self,ying2018hierarchical,zhang2018end}.

\subsection{Computational Tasks}
Computational tasks related to graphs can generally be classified into two primary categories: node-level and graph-level tasks. At the node level, the main tasks involve classifying nodes~\cite{ju2023zero}, ranking nodes~\cite{agarwal2006ranking}, clustering nodes~\cite{yi2024redundancy} and predicting links between nodes~\cite{zhang2018link}. On the other hand, tasks at the graph level primarily include classifying entire graphs~\cite{ju2022tgnn,ju2024hypergraph}, matching different graphs~\cite{caetano2009learning} and generating new graphs~\cite{guo2022systematic}. Here we introduce three representative computational tasks on the graph.

\smallskip\textbf{Node Classification.}
Given graph $\mathcal{G}=\{\mathcal{V},\mathcal{E}\}$, with a set of labeled nodes denoted as $\mathcal{V}_{\text{L}}\subset \mathcal{V}$ and unlabeled nodes set denoted as $\mathcal{V}_{\text{U}}=\mathcal{V}\setminus\mathcal{V}_{\text{L}}$. We assume each node $i\in\mathcal{V}_{\text{L}}$ is associated with a label $y_i$. Node classification aims to learn a model using $\mathcal{G}$ and its available label set to predict the labels of the unlabeled nodes in $\mathcal{V}_{\text{U}}$.

\smallskip\textbf{Link Prediction.}
Given graph $\mathcal{G}=\{\mathcal{V},\mathcal{E}\}$, with a set of known edges $\mathcal{E}_{\text{K}}$. We assume the existence of unobserved links $\mathcal{E}_{\text{U}}=\mathcal{E}\setminus\mathcal{E}_{\text{K}}$ between nodes. Link prediction aims to learn a model from $\mathcal{G}$ to predict whether a link exists between nodes $v_i$ and $v_j$, where $(v_i,v_j)\notin \mathcal{E}_{\text{K}}$.

\smallskip\textbf{Graph Classification.}
Beyond tasks focused on individual nodes, graph classification operates at the level of entire graphs. Given a training graph dataset $\mathcal{D}=\{(\mathcal{G}_i, y_i)\}_{i=1}^{|\mathcal{D}|}$ with multiple graphs, we assume each graph instance belongs to a certain class, where $y_i$ is the label of graph $\mathcal{G}_i$ and $|\mathcal{D}|$ represents the total number of graphs in the training set. Graph classification aims to learn a model from the dataset $\mathcal{D}$ to predict the labels of unseen test graphs.

Based on the basic concept of graphs, more details about GNN applications in real-world practical scenarios can be found in the following sections.

%% file: 5_imbalance.tex
\section{Imbalance}
\begin{figure}
\centering
\includegraphics[width=\linewidth]{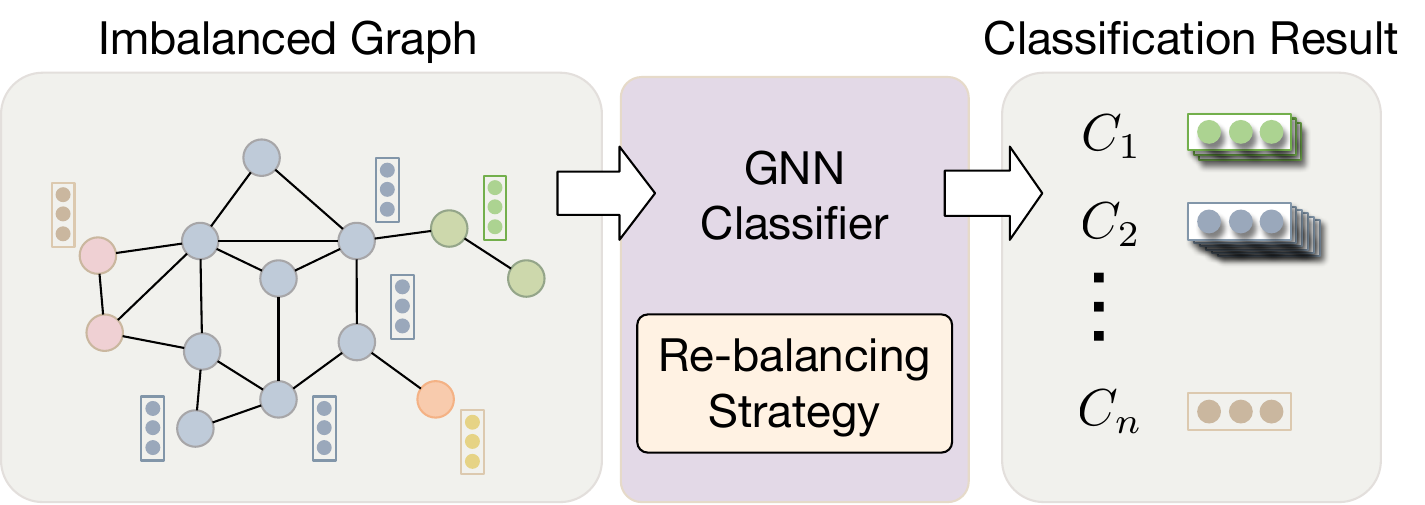}
\caption{Illustration of the data imbalanced problem. The labels assigned to nodes or graphs that obtained from real-world data sources always suffer from severe class imbalance issue brought by the long-tail distribution of samples. The challenge calls for various applicable re-balancing strategies to train robust and reliable GNNs.}
\label{fig:imbalance}
\end{figure}

In real-world GNN applications, data imbalance poses a major challenge, often showing significant disparities in class instance counts. This is common in tasks like fraud detection~\cite{liu2021pick} and anomaly detection~\cite{liu2023qtiah}. The dominance of majority classes leads to minority classes being underrepresented, harming overall performance. Formally, let $\{C_1, C_2, ..., C_K\}$ denote $K$ classes with $|C_k|$ samples in class $k$, satisfying $|C_1|\geq |C_2|\geq\dots\geq |C_K|$. 
The goal is to train a GNN classifier that performs well on both majority (e.g., $C_1$) and minority (e.g., $C_K$) classes. 
In other words, we expect that the trained GNN classifier \( \mathcal{F}^* \) can maximize the utility, such as the accuracy (ACC) for each class, i.e., 
\begin{equation}
    \mathcal{F}^* = \underset{\mathcal{F}}{\arg\max} ~ \{\text{ACC}(\mathcal{F}(C_k))\}_{k=1}^K. 
\end{equation}
To tackle this, existing methods are broadly grouped into three categories: \emph{re-balancing}, \emph{augmentation-based}, and \emph{module improvement} methods. Figure~\ref{fig:imbalance} illustrates the concept, and Table~\ref{tab:imbalance} summarizes representative works. 
Next, we delve into each strategy, offering a comprehensive overview.

\subsection{Re-balancing Approaches}

Re-balancing approaches aim at addressing the problem of uneven distribution of training samples across different classes, including two main categories of methods: \emph{re-sampling}, and \emph{cost-sensitive learning}.

\smallskip\textbf{Re-sampling (RS).} RS adjusts the selection of samples during training. 
% Typically, samples are chosen at random, leading to a bias towards the majority classes. Re-sampling addresses this by ensuring an equitable distribution of samples from all classes in the training process. 
Standard RS techniques involve either replicating samples in the minority class or reducing samples in the majority class. However, in cases of severe imbalance, they can either cause overfitting or weaken performance, respectively. 
Therefore, recent studies mainly aim to synthesize ($\mathcal{B}_{\text{SYN}}$) minority samples, or partition ($\mathcal{B}_{\text{PAR}}$) the majority samples to balance ($\mathcal{B}$) the classes, formulated as: 
\begin{equation}
    \mathcal{B}(C_1,\ldots,C_n) = \{\mathcal{B}_{\text{SYN}}(C_{\text{minority}}),\mathcal{B}_{\text{PAR}}(C_{\text{majority}})\}, 
\end{equation}
where $C_{\text{minority}}$ and $C_{\text{majority}}$ contain samples from the majority and minority classes, respectively. 
GraphSMOTE \cite{zhao2021graphsmote} employs synthetic minority oversampling within the embedding space to increase the representation of minority classes. Moreover, it integrates an edge generator to create new connections between synthesized samples and existing ones, which can produce dependable relational data among samples. 
To improve sample quality, ImGAGN~\cite{qu2021imgagn} introduces a generative adversarial graph network that generates synthetic minority nodes and uses a GCN-based discriminator to distinguish real nodes. Despite effectiveness, they still struggle with neighbor memorization under severe imbalance. GraphENS~\cite{park2022graphens} addresses this by generating minority nodes along with their one-hop neighbors, synthesizing complete ego networks. 
SNS~\cite{gao2023semantic} tackles imbalanced heterogeneous networks by adaptively selecting neighbors and enriching the network with synthetic nodes. 
DataDec~\cite{zhang2023sparsity} proposes a dynamic sparsity framework—data decantation—that selects informative samples via gradient score ranking. 
Recently, C$^3$GNN~\cite{ju2025cluster} proposes a cluster-guided contrastive framework, which alleviates imbalance by dividing majority classes into balanced subclasses and enriching them via mixup.

\smallskip\textbf{Cost-sensitive Learning (CSL).} 
CSL adjusts training loss for various classes to tackle the imbalances in training. A widely used method involves applying the frequencies of labels from the training data to adjust the weights of the loss function. As a variation of this technique, class-balanced loss ($\ell_{\text{CB}}$) \cite{cui2019class} involves scaling the loss of various classes according to the inverse of the effective number of samples in each class, formulated as: 
\begin{equation}
    \ell_{\text{CB}} = - \sum_{i=1}^{|\mathcal{V}_{\text{L}}|} \frac{1-\beta}{1-\beta^{|C_{y_i}|}}\log(\ell_{\text{ERM}, i}),
\end{equation}
where 
% $n_y$ stands for the number of samples of class $y$ and 
$\ell_{\text{ERM}, i}$ is the empirical risk of each labeled sample and $\beta$ is a hyperparmeter.
However, this direct approach may not always be the best solution because it does not consider the graph topology. To address this, ReNode \cite{chen2021topology} focuses on both the quantity and topology imbalances in nodes by examining the shift in node influence and adaptively adjusting the weight of labeled nodes according to their relative positions to class boundaries. Similarly, TAM \cite{song2022tam} also utilizes topological information, comparing each node's connectivity pattern to the average pattern of its class and adaptively modifying the margin based on that.

\begin{table}[t]
% \caption{Overview of methods for learning from imbalanced graphs, categorized into three main types: re-balancing, augmentation-based, and module improvement methods. In this table, ``CSL" denotes cost-sensitive learning, ``TL" symbolizes transfer learning, ``IA" stands for information augmentation, ``RL" represents representation learning, and ``CT" refers to classifier training.}
\caption{Overview of methods for learning from imbalanced graphs, categorized into three main types: re-balancing, augmentation-based, and module improvement methods. 
% (``CSL": cost-sensitive learning, ``TL": transfer learning, ``IA": information augmentation, ``RL": representation learning, ``CT": classifier training)
}
\label{tab:imbalance}
\centering
\setlength{\tabcolsep}{3pt}
\resizebox{1\columnwidth}{!}{
\begin{tabular}{lcccccccc}
\toprule
\multirow{2}{*}{Method} & \multirow{2}{*}{Task Type} & \multicolumn{2}{c}{Re-balancing} & \multicolumn{2}{c}{Augmentation} & \multicolumn{3}{c}{Module Improvement} \\
\cmidrule{3-4} \cmidrule{5-6} \cmidrule{7-9}
 & &RS &CSL &TL &IA &RL &CT & ME \\
 \midrule
GraphSMOTE \cite{zhao2021graphsmote} & Node-level & \checkmark & & & & & & \\
ImGAGN \cite{qu2021imgagn} & Node-level & \checkmark & & & & & & \\
GraphENS \cite{park2022graphens} & Node-level & \checkmark & & & & & & \\
SNS \cite{gao2023semantic} & Node-level & \checkmark & & & & \checkmark & & \\
C$^3$GNN \cite{ju2025cluster} & Graph-level & \checkmark & & & & \checkmark & & \\
DataDec \cite{zhang2023sparsity} & Graph-level & \checkmark & & & & & & \\
% CB Loss \cite{cui2019class} & - & &\checkmark & & & & & \\
ReNode \cite{chen2021topology} & Node-level & &\checkmark & & & & & \\
TAM \cite{song2022tam}  & Node-level & &\checkmark & & & & & \\
SOLTGNN \cite{liu2022size} & Graph-level & & &\checkmark & & & & \\
RAHNet \cite{mao2023rahnet} & Graph-level & & &\checkmark & &\checkmark &\checkmark & \\
GNN-INCM \cite{huang2022graph} & Node-level & & &\checkmark & & & &\checkmark \\
LTE4G \cite{yun2022lte4g} & Node-level & & &\checkmark & & & &\checkmark \\
GraphMixup \cite{wu2022graphmixup} & Node-level & & & &\checkmark & & & \\
G$^2$GNN \cite{wang2022imbalanced} & Graph-level & & & &\checkmark &\checkmark & & \\
CM-GCL \cite{qian2022co} & Node-level & &\checkmark & &\checkmark & & & \\
INS-GNN \cite{juan2023ins}  & Node-level & & & & &\checkmark & & \\
ImGCL \cite{zeng2023imgcl} & Node-level &\checkmark & & & &\checkmark & & \\
GNN-CL \cite{li2024graph} & Node-level & & & & &\checkmark & & \\
GraphDIVE \cite{hu2022graphdive}  & Graph-level & & & & & & &\checkmark \\
CoMe \cite{yi2023towards} & Graph-level & &\checkmark & & &\checkmark & &\checkmark \\
PASTEL \cite{sun2022position} & Node-level & & & &\checkmark &\checkmark & & \\
QTIAH-GNN \cite{liu2023qtiah} & Node-level & & \checkmark & & &\checkmark & & \\
\bottomrule
\end{tabular}
}
\end{table}

\subsection{Augmentation-based Approaches}

Augmentation-based approaches aim to enhance model training with extra information, boosting performance in imbalanced learning scenarios. This approach includes two techniques: \emph{information augmentation} and \emph{transfer learning}.

\smallskip\textbf{Information Augmentation (IA).} 
% This category of method aims to introduce additional knowledge into model training. GraphMixup \cite{wu2022graphmixup} enhances graph data augmentation to address imbalances in several ways. It implements Semantic Feature Mixup at the semantic level, introduces Contextual Edge Mixup for local and global structure insights, and develops a reinforcement mechanism to adaptively set the upsampling ratio for each minority class. Instead of directly augmenting samples, G$^2$GNN \cite{wang2022imbalanced} constructs a graph-of-graph based on kernel similarity, gaining additional global supervision from neighboring graphs and local supervision directly from the graphs themselves. Moreover, CM-GCL \cite{qian2022co} leverages multi-modality data and develops a co-modality framework that automatically generates contrastive pairs. The overall framework is optimized by both inter-modal and intra-modal graph contrastive losses.
IA introduces additional knowledge into model training. 
IA techniques includes generating informative augmented views through transformations $\mathcal{T}_{\text{IA}}=\mathcal{T}_{\text{IA}}(\mathcal{G})=\{ \tilde{G}_1, \tilde{G}_2, \dots \}.$ 
These augmented data provide complementary learning signals that help alleviate class imbalance. The overall training loss typically consists of two components:
\begin{equation}
\ell = \ell_{\text{ERM}}(\mathcal{F}(\mathcal{G}), \mathbf{Y}) + \lambda \ell_{\text{AUX}}(\mathcal{T}_{\text{IA}}),
\end{equation}
where \(\ell_{\text{ERM}}\) is the standard empirical risk, e.g., cross-entropy loss, loss on the original graphs, and \(\ell_{\text{AUX}}\) is an auxiliary loss that depends on the augmented data \(\mathcal{T}_{\text{IA}}\).
For example, GraphMixup~\cite{wu2022graphmixup} enhances graph augmentation via semantic feature mixup and contextual edge mixup for local and global structure, and a reinforcement mechanism that adaptively sets the upsampling ratio for each minority class. 
However, this augmentation lacks effective supervision and discriminative power. 
Instead of augmenting samples directly, G$^2$GNN~\cite{wang2022imbalanced} builds a graph-of-graphs using kernel similarity to gain global supervision from neighbors and local supervision from the graphs themselves. CM-GCL~\cite{qian2022co} exploits multi-modal data through a co-modality framework that generates contrastive pairs, and applies inter- and intra-modal graph contrastive losses for optimization.

\smallskip\textbf{Transfer Learning (TL).} 
% TL aims to apply knowledge gained from one domain (like specific datasets or classes) to improve model training in another domain. In the context of imbalanced learning in graphs, there are two main approaches: majority-to-minority knowledge transfer and knowledge distillation. 
% The goal of transferring knowledge from majority to minority classes is to leverage the knowledge gained from the majority categories to improve the model's effectiveness in predicting the minority categories.
% For example, SOLTGNN \cite{liu2022size} exploits the co-occurrence substructures on majority graphs from both node- and subgraph-levels, and employs a relevance prediction function to identify and transfer patterns from majority graphs to minority graphs. RAHNet \cite{mao2023rahnet} proposes a retrieval augmented branch to retrieve the most relevant graphs and introduce new knowledge to the minority classes, thereby enhancing their representational ability for the minority classes. Knowledge distillation represents another approach within the realm of transfer learning. GNN-INCM \cite{huang2022graph} presents a knowledge distillation module that focuses on hard samples. This module enables the concurrent training of several GNN models by employing distribution and triplet alignment losses. LTE4G \cite{yun2022lte4g} applies knowledge distillation to create two student models: one tailored for nodes in the majority classes, while the other dedicated to nodes in the minority classes, each responsible for classifying nodes in their respective classes.
TL aims to apply knowledge from one domain (e.g., specific datasets or classes) to enhance model training in another. In graph imbalanced learning, two main TL approaches are majority-to-minority transfer and knowledge distillation. The former leverages patterns from majority classes to improve prediction on minority classes. For instance, SOLTGNN~\cite{liu2022size} captures co-occurrence substructures from majority graphs at both node- and subgraph-levels and uses a relevance predictor to transfer them to minority graphs. RAHNet~\cite{mao2023rahnet} introduces a retrieval-augmented branch that retrieves informative graphs to enrich minority class representations. As another direction, knowledge distillation enables knowledge transfer via teacher-student training. GNN-INCM~\cite{huang2022graph} incorporates a distillation module focusing on hard samples using distribution and triplet alignment losses. LTE4G~\cite{yun2022lte4g} employs two student models: one for majority nodes and another for minority nodes, each specialized in classifying their respective classes.

\subsection{Module Improvement Approaches}
Research in this area focuses on enhancing network modules in imbalanced learning, encompassing \emph{representation learning}, \emph{classifier training}, and \emph{model ensemble}. 

% \smallskip\textbf{Representation Learning.} INS-GNN \cite{juan2023ins}  initially employs self-supervised learning to pre-train the model, and then leverages self-training to assign pseudo-labels to unlabeled nodes. While contrastive learning is typically used to improve representation, ImGCL \cite{zeng2023imgcl} identifies a limitation in current graph contrastive methods' ability to discriminate imbalanced nodes. To address this, ImGCL introduces a node centrality-based progressively balanced sampling method, aimed at better maintaining the intrinsic structure of graphs. In addition to self-supervised learning, GNN-CL \cite{li2024graph} employs metric learning, focusing on distance-based losses to learn an embedding space with better discrimination capabilities. Specifically, it introduces Neighbor-based Triplet Loss to differentiate among samples associated with the minority class within the feature space by modifying the distances between nodes.
\smallskip\textbf{Representation Learning (RL).} 
While contrastive learning is commonly used to enhance representation~\cite{luo2022clear,ju2023unsupervised}, ImGCL~\cite{zeng2023imgcl} identifies a limitation in current graph contrastive methods’ ability to discriminate imbalanced nodes. To address this, ImGCL introduces a node centrality-based progressively balanced sampling method to better preserve the intrinsic graph structure. But no extra supervision is introduced. 
INS-GNN~\cite{juan2023ins} first employs self-supervised learning to pre-train the model, then uses self-training to assign pseudo-labels to unlabeled nodes. Besides self-supervised learning, GNN-CL~\cite{li2024graph} applies metric learning, focusing on distance-based losses to learn an embedding space with improved discrimination. Specifically, it proposes neighbor-based triplet loss to distinguish minority-class samples by adjusting node distances in the feature space.

% \smallskip\textbf{Classifier Training.} For better classifier training, RAHNet \cite{mao2023rahnet} jointly learns a balanced feature extractor and an unbiased classifier with decoupling training. Typically, imbalanced class distribution leads to larger weight norms for majority classes, which can bias the classifier towards these dominant classes. RAHNet tackles this issue by regularizing the classifier's weights while keeping the trained feature extractor fixed. 
\smallskip\textbf{Classifier Training (CT).} Imbalance causes larger weight norms for majority classes, biasing the classifier toward dominant ones. For better training, RAHNet~\cite{mao2023rahnet} jointly learns a balanced feature extractor and unbiased classifier via decoupling. RAHNet mitigates classifier bias by regularizing classifier weights while keeping the extractor fixed. 

\smallskip\textbf{Model Ensemble (ME).} 
ME typically employs a multi-expert learning framework to conduct graph learning from diverse perspectives, thereby enhancing model performance. 
A typical ME method takes the form: 
\begin{equation}
\hat{y} = \sum_{i=1}^{K} w_i \cdot \mathcal{F}_i(\mathcal{G}_i), \quad \sum_{i=1}^{K} w_i = 1,
\end{equation}
where each \(\mathcal{F}_i\) denotes the prediction from expert \(i\), trained on a specific view, and \(w_i\) is its gating weight. Diversity is introduced by varying inputs or architectures, and the weights $w_i$ may be static or learned dynamically based on input.
GraphDIVE~\cite{hu2022graphdive} focuses on extracting varied representations at both node and graph levels. To ensure the diversity of experts, each expert receives a specific graph representation view, which forms the final prediction through aggregation. 
Building upon the multi-expert framework, CoMe~\cite{yi2023towards} combines various expert models using dynamic gating functions, enhancing the overall diversity of the training network. Moreover, it performs knowledge distillation in a disentangled manner among experts, motivating them to learn extra knowledge from each other.

\subsection{Discussion}

% Although significant advancements have been made in graph imbalanced learning, the focus has predominantly been on label imbalance, with less attention given to structural imbalances within graphs. Sun et al. \cite{sun2022position} examine the distribution of supervision information from a comprehensive global perspective, focusing on the challenges of under-reaching and over-squashing. To tackle these challenges, they introduce a position-aware structure learning module that optimizes information propagation paths, addressing topology imbalance directly. QTIAH-GNN \cite{liu2023qtiah} also focuses on addressing the topology imbalance issue in GNN. It introduces a multi-level, label-aware neighbor selection mechanism that aims to identify and sample neighbors similar to a given central node while effectively excluding nodes of dissimilar classes. However, further research is needed to deepen our understanding of topology imbalance and develop effective solutions for it.
Although significant advancements have been made in graph imbalanced learning, most work focuses on label imbalance, with less attention given to structural imbalances within graphs. 
Sun et al.~\cite{sun2022position} analyze supervision distribution globally, addressing under-reaching and over-squashing via position-aware structure learning that optimizes information propagation paths, addressing topology imbalance directly. Similarly, QTIAH-GNN~\cite{liu2023qtiah} tackles topology imbalance with a multi-level, label-aware neighbor selection mechanism that aims to identify and sample neighbors similar to a given central node while effectively excluding nodes of dissimilar classes. However, deeper understanding and solutions for topology imbalance remain needed.

%% file: 6.1_label_noise.tex
\section{Noise}

\begin{figure}
\centering
\includegraphics[width=\linewidth]{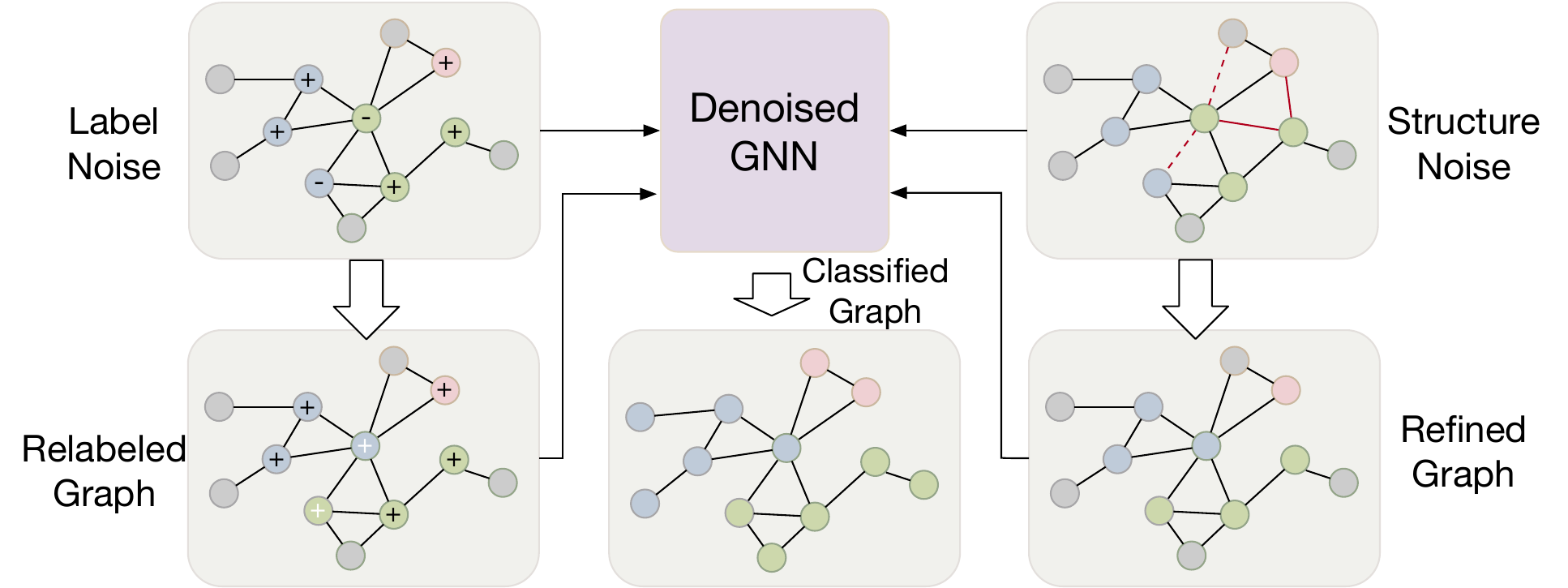}
\caption{Illustration of GNNs under the impact of label and structural noise. Inevitable label errors require the GNN model to accurately identify mislabeled samples, while the fake or absent edges between nodes require the model to reconstruct the ground-truth adjacency.}
\label{fig:noise}
\end{figure}

In addition to the challenge of class imbalance, the presence of data noise within the graph is a widespread issue in real-world scenarios~\cite{jin2025systematic}. There are three common types of noise: \emph{label noise}, \emph{structural noise} and \emph{attribute noise}. Figure~\ref{fig:noise} illustrates the fundamental principles of GNNs in addressing three noises. Next, we discuss these three aspects in detail.

\subsection{Label Noise}

This section introduces graph learning with label noise, covering both node-level \cite{dai2021nrgnn,qian2023robust,li2023graphcleaner,du2023noise} and graph-level \cite{nt2019learning,yin2023omg,yin2024sport} classification. The former is more extensively studied, and our focus is on node-level classification. The goal is to train a GNN classifier $\mathcal{F}^{*}$ that is robust to label noise. The model is trained on a graph $\mathcal{G}$ where some nodes have noisy labels $\mathbf{Y}^{\text{noisy}}$, and outputs predicted labels $\mathbf{Y}_{\text{U}}$ for the unlabeled nodes as: 
% $\hat{y}_{j} =  \mathcal{F}^{*}(\mathcal{G})$, $j \in \mathcal{V}_{\text{U}}$, where
\begin{align*}
    \mathbf{Y}_{\text{U}} = & \mathcal{F}^{*}(\mathcal{G}), ~
    \mathcal{F}^{*} = \underset{\mathcal{F}}{\arg \min }~ \ell(\mathcal{F}(\mathcal{G}),\mathbf{Y}^{\text{noisy}}).
\end{align*}
The expectation is that the trained model effectively mitigates the adverse impact of noisy labels on the predictions $\mathbf{Y}_{\text{U}}$ for the unlabeled nodes. 
Studies addressing scenarios with noisy labels on graphs can be categorized into two groups: \emph{loss correction} and \emph{label correction}. Table \ref{table:noisylabelstructure}(a) presents the overview of these methodologies. In the subsequent sections, we delve into detailed introductions. 

\subsubsection{Loss Correction Approaches}

Loss correction (LoC) approaches \cite{dai2021nrgnn,qian2023robust} aim to rectify the impact of label noise on risk minimization by adapting the training loss. Various commonly employed techniques include \emph{loss regularization}, \emph{sample reweighting},  \emph{adversarial attack and defense}, and \emph{guidance from a mislabeled transition matrix}. 

\smallskip\textbf{Loss Regularization.} Loss regularization methods introduce additional information to alleviate the impact of noisy labels, where the total loss can be formulated as:
\begin{equation}
    \ell = \ell_{\text{ERM}}(\mathcal{F}(\mathcal{G}),\mathbf{Y}^{\text{noisy}}) + \lambda \ell_{\text{REG}},
\end{equation}
where $\ell_{\text{ERM}}$ is the empirical risk and $\ell_{\text{REG}}$ is the regularization term with additional information. 
By incorporating such supplementary information, loss regularization enhances the model's robustness to noisy labels during training. 
% For example, NRGNN~\cite{dai2021nrgnn} connects unlabeled and labeled nodes via high feature similarity, introducing structure reconstruction loss to adjust training. It also incorporates high-confidence pseudo-labels to enhance supervision and further reduce label noise. 
For example, NRGNN \cite{dai2021nrgnn} suggests connecting unlabeled nodes with labeled nodes by exploring high feature similarity, where the structure reconstruction loss is introduced to adjust the training loss. This approach also incorporates pseudo-label information with high prediction confidence to enhance supervision and further mitigate the impact of label noise. However, it overlooks the potential negative propagation caused by such connections. 
DND-Net~\cite{ding2024divide} proposes a simple yet robust framework that decouples propagation and transformation to prevent noise spreading, integrating reliable pseudo-labeling with neighbor-aware uncertainty reweighting. 
From another perspective, PIGNN~\cite{du2023noise} focuses on pair-wise interaction (PI) between nodes to support noise-tolerant GNN learning, showing that PI introduces less noise than point-wise methods. 
Contrastive learning (CL)~\cite{chen2020simple,luo2023self,ju2025gps,ju2024towards} leverages comparisons between similar and dissimilar pairs to learn robust representations. 
CR-GNN~\cite{li2024contrastive} uses an unsupervised neighbor contrastive loss with a dynamic cross-entropy loss, selecting nodes with consistent predictions as reliable to prevent overfitting to noisy labels.

\smallskip\textbf{Sample Reweighting.} 
% Leveraging the memorization effect, training on small-loss samples emerges as a particularly promising approach for mitigating the challenges posed by noisy labels \cite{yu2019does}. In the context of sample reweighting, the strategy involves down-weighting large-loss samples in the training loss, thereby augmenting the supervision capability associated with clean labels. The corresponding loss can be unified into: 
Leveraging the memorization effect, training on small-loss samples is a promising approach to mitigate challenges posed by noisy labels~\cite{yu2019does}. In sample reweighting, the strategy down-weights large-loss samples in the training loss, thereby enhancing the supervision from clean labels. The corresponding loss can be unified into: 
\begin{equation}
\ell =  \sum_{i=1}^{|\mathcal{V}_{\text{L}}|} \omega_{i} \cdot \ell_{\text{ERM}, i}, \quad \sum_{i=1}^{|\mathcal{V}_{\text{L}}|} w_i = 1,
\end{equation}
where $\omega_{i}$ is the specific-defined weight and $\ell_{\text{ERM}, i}$ is the empirical risk of each labeled node. 
Drawing inspiration from the edge predictor \cite{dai2021nrgnn}, RTGNN \cite{qian2023robust} integrates both sample reweighting and loss regularization techniques. Utilizing the small-loss principle (SLP), RTGNN filters out clean labels and diminishes the influence of noisy labels in the training process. Additionally, 
% RTGNN introduces self-reinforcement and consistency regularization as auxiliary forms of supervision, aiming to enhance the robustness of the model. 
RTGNN introduces internal strengthening and coherence regularization as supplementary forms of supervision, aiming to enhance the model's robustness.

\smallskip\textbf{Adversarial Attack and Defense.} 
% Akin to noisy labels, adversarial label-flipping attacks involve strategically manipulating labels to mislead a model during training, but with the intent of deliberately introducing misclassifications for adversarial purposes. 
% CP \cite{zhang2020adversarial} tackles label-flipping attacks by developing an attack model called LafAK. This model is built upon an approximated closed form of GNN and employs a continualization strategy for the non-differentiable objective. A defense framework is suggested, incorporating self-supervised task that preserves community property as a regularization strategy to alleviate overfitting. 
Akin to noisy labels, adversarial label-flipping attacks strategically manipulate labels to mislead a model during training, deliberately introducing misclassifications for adversarial purposes. CP \cite{zhang2020adversarial} develops an attack model LafAK, based on an approximated closed form of GNNs, to simulate label-flipping attacks. They also propose a defense framework that uses a self-supervised task to preserve community properties as regularization, thus improving robustness and mitigating overfitting.

\smallskip\textbf{Mislabeled Transition Matrix (MTM).} 
MTM is instrumental in characterizing how nodes in different classes are mislabeled, effectively capturing the underlying pattern of noise formation. By utilizing this matrix, it guides the training process when dealing with noisy labels. GraphCleaner \cite{li2023graphcleaner} first leverages the validation set to learn MTM and then uses the estimated MTM as a synthetic mislabel dataset generator to train the noise detector.

\subsubsection{Label Correction Approaches}

Label correction (LaC) approaches \cite{xia2023gnn,yuan2023learning} offer a more intuitive solution by identifying nodes with potentially incorrect labels and correcting them to ensure reliable training. The detect-and-correct procedure can be formulated as:
\begin{equation}
    \mathbf{Y}^{\text{correct}} = \mathcal{C}_{\text{N}}(\mathcal{G}, \mathbf{Y}^{\text{noise}}), ~
    \mathbf{Y}^{\text{noise}} = \mathcal{D}_{\text{N}}(\mathcal{G}, \mathbf{Y}^{\text{noisy}}), 
\end{equation}
where $\mathcal{C}_{\text{N}}$ and $\mathcal{D}_{\text{N}}$ are tailored noise corrector and detector, respectively. Then the training loss is built upon the corrected $\mathbf{Y}^{\text{correct}}$. 
Common correction techniques include \emph{label propagation}, \emph{neighbor voting}, et al. These techniques contribute to refining the accuracy of labels in the training data, ultimately enhancing the robustness of the model against the impact of noisy labels during the learning process.

\smallskip\textbf{Label Propagation (LP).} 
In the context of the homophily assumption, LP disseminates clean labels through the graph by leveraging node similarity, effectively correcting potentially mislabeled nodes. 
For example, UnionNET~\cite{li2021unified} uses similarities of learned node representations as attention weights for label aggregation, then employs aggregated class probabilities to weight training samples and guide label correction, jointly optimizing the network. However, it fails to distinguish the reliability of samples.
GNN Cleaner~\cite{xia2023gnn} propagates clean labels through graph structure, generating pseudo-labels filtered by agreement with given labels, and then applies a learnable correction scheme supervised by reliable labels. 
Coding rate reduction (CRR)~\cite{ma2007segmentation} promotes semantically rich representations, extended by ERASE~\cite{chen2024erase} via decoupled propagation that combines placeholder and refined labels with fault-tolerant adjustments. 
Additionally, LP4GLN~\cite{cheng2023label} tackles noisy labels in heterogeneous graphs by reconstructing homophily-restored graphs, iteratively selecting high-confidence labels through LP.

\smallskip\textbf{Neighbor Voting (NV).} 
% Neighbor voting determines the corrected label for a node by considering the majority label among its neighboring nodes. 
% This approach assumes that neighboring nodes are likely to be similar, 
% emphasizing the influence of local neighborhood information in correcting potentially noisy labels. 
% CGNN \cite{yuan2023learning} is a product of combining both loss correction and label correction. It utilizes graph contrastive learning as a regularization term, avoiding the use of label information to prevent overfitting to noisy labels. It filters noisy labels through the consistency between the predictive/annotated labels of nodes and neighbors to rationally leverage label information. For the filtered noisy nodes, it corrects labels using a neighbor voting mechanism. 
NV determines the corrected label for a node by considering the majority label among its neighbors. This approach assumes neighboring nodes are likely to be similar, emphasizing the role of local structure in correcting noisy labels. 
CGNN~\cite{yuan2023learning} combines Loc and Lac, employing graph CL as a regularization to avoid overfitting. It filters noisy labels based on consistency between predicted/annotated labels and neighbors, then corrects filtered noisy nodes via neighbor voting.

\begin{table}[t]
    \caption{Overview of methods on graphs against three types of noises. 
    (``PLL'': pseudo-label learning, ``SR'': self-reinforcement, ``HR'': homophily reconstruction, ``ES'': edge-level sampling, ``NS'': node-level sampling)}
    \label{table:noisylabelstructure}
    \centering
    {\bf (a).} Label Noise 
    \vspace{2mm}

    \setlength{\tabcolsep}{2pt}
    \resizebox{1\columnwidth}{!}{
    \begin{tabular}{lccc}
        \toprule
        Method & Data Type & Core Idea & Implementation \& Details \\
        \midrule
        NRGNN \cite{dai2021nrgnn} & Node-level & LoC & Edge Connection, PLL  \\
        DND-Net \cite{ding2024divide} & Node-level & LoC & Decoupling, PLL \\
        PIGNN \cite{du2023noise} & Node-level & LoC & PI \\
        CR-GNN \cite{li2024contrastive} & Node-level & LoC & CL, Sample Selection \\
        CP \cite{zhang2020adversarial}  & Node-level & LoC & Attack and Defense \\
        RTGNN \cite{qian2023robust} & Node-level & LoC & SLP, SR \\
        GraphCleaner \cite{li2023graphcleaner} & Node-level & LoC & MTM, Neighbor Agreement \\
        UnionNET \cite{li2021unified}& Node-level & LaC & Label Aggregation \\
        GNN Cleaner \cite{xia2023gnn} &  Node-level & LaC  & LP \\
        ERASE \cite{chen2024erase} &  Node-level & LaC  &  CRR, LP \\
        CGNN \cite{yuan2023learning} & Node-level & LoC \& LaC  & CL, NV\\
        LP4GLN \cite{cheng2023label} & Node-level & LaC &  HR, LP \\
        D-GNN \cite{nt2019learning} &  Graph-level & LoC  & MTM, Backward LoC \\
        OMG \cite{yin2023omg} & Graph-level & LoC \& LaC & Coupled Mixup, CL \\
        \bottomrule
    \end{tabular}
    }

    \vspace{2mm}
    {\bf (b).} Structure Noise 
    \vspace{2mm}

    \setlength{\tabcolsep}{2pt}
    \resizebox{1\columnwidth}{!}{
    \begin{tabular}{lccc}
        \toprule
        Method & Type & Post-processing & Graph Regularization \\ \midrule
        GRCN \cite{yu2021graph} & KML & $k$NN & Sparsification \\
        GNNGuard \cite{zhang2020gnnguard} & KML & $\epsilon$NN & Sparsification \\
        GDC \cite{gasteiger2019diffusion} & KML & $k$NN,$\epsilon$NN & Sparsification \\
        GLCN  \cite{jiang2019semi} & NML & $k$NN & Sparsification \\
        IDGL  \cite{chen2020iterative} & NML & $\epsilon$NN & Sparsification, Smoothness \\
        SLAPS  \cite{fatemi2021slaps} & NML & $k$NN & Sparsification \\
        DropEdge \cite{rong2020dropedge} & ES & $\epsilon$NN & Sparsification \\
        DropCONN \cite{chen2020enhancing} & ES & - & Smoothness \\
        FastGCN  \cite{chen2018fastgcn}& NS & - & - \\
        PTDNet \cite{luo2021learning}& ES & - & Sparsification,  Smoothness \\
        NeuralSparse \cite{zheng2020robust} & ES & $k$NN & Sparsification \\
        % GraphSAGE \cite{hamilton2017inductive}& Node-level sampling & $k$NN & Sparsification \\
        TO-GCN \cite{yang2019topology} & DO & - & Smoothness \\
        PRO-GNN \cite{jin2020graph}& DO & - & Sparsification,  Smoothness \\
        % DHGR  \cite{bi2022make}& DO & - & Smoothness \\
        % GLNN \cite{gao2020exploring}& DO & - & Sparsification,  Smoothness \\ 
        Xu et al.~\cite{xu2019topology} & DO & - & Attack and Defense \\
        Gosch et al.~\cite{gosch2023adversarial} & DO & - & Attack and Defense \\
        PTA \cite{dong2021equivalence}  & DO & - & Decoupling \\ 
        RLP \cite{he2022structural}  & DO & - & Decoupling \\ 
        PAMT \cite{chen2024pamt}  & DO & - & Decoupling \\
        \bottomrule
    \end{tabular}
    }

    \vspace{2mm}
    {\bf (c). Attribute Noise}
    \vspace{2mm}
    
    \setlength{\tabcolsep}{2pt}
    \resizebox{1\columnwidth}{!}{
    \begin{tabular}{lccc}
        \toprule
        Method & Type  & Implementation \& Details \\
        \midrule
        Nettack~\cite{zugner2018adversarial} & Attack and Defense  & Incremental Computations  \\
        BVAT~\cite{deng2023batch} & Attack and Defense & Virtual Adversarial Perturbations  \\
        GCORN~\cite{abbahaddou2024bounding} & Defense  & Orthonormal Weight Matrix \\
        MQE~\cite{li2024noise} & Loss Refinement  & Multi-hop Propagation \\
        BRGCL~\cite{wang2022bayesian} & Loss Refinement  & Contrastive Learning \\
        \bottomrule
    \end{tabular}
    }
\end{table}

%% file: 6.2_structure_noise.tex
\subsection{Structure Noise}
Structure noise in GNNs refers to the presence of irrelevant or noisy information in the graph structure that can negatively impact the performance of the GNN model \cite{fox2019robust}. 
GNNs are highly susceptible to structure noise since errors can propagate throughout the graph due to the message-passing mechanism \cite{gilmer2017neural}. 
Therefore, the quality of the input graph structure is critical to achieving optimal GNN performance \cite{de2013influence}.
The mainstream methods of tackling structure noise are \emph{graph structure learning} and \emph{direct optimization}. 
The former focuses on optimizing the graph structure before carrying out downstream tasks and can be further categorized into \emph{metric learning} and \emph{sampling-based} approaches. 
They refine the noisy adjacency matrix and node representations by: 
\begin{equation}
   \mathbf{A^*} = \mathcal{F}_{\text{ADJ}}(\mathbf{A}^{\text{noisy}}, \mathbf{X}) , ~\mathbf{Z}^* = \mathcal{F}_{\text{REP}}(\mathbf{X}, \mathbf{A^*}), 
\end{equation}
where $\mathcal{F}_{\text{ADJ}}$ and $\mathcal{F}_{\text{REP}}$ are the structure learner and representation learner, respectively. 
The latter mitigates the impact of structural noise by directly incorporating tailored regularization terms ($\ell_{\text{REG}}$), improving the GNN architecture ($\bar{\mathcal{F}}$), or refining the supervision labels ($\bar{\mathbf{Y}}$) to optimize the training loss, where the loss can be unified as: 
\begin{equation}
   \ell = \ell_{\text{ERM}}(\bar{\mathcal{F}}(\mathbf{A}^{\text{noisy}}, \mathbf{X}), \bar{\mathbf{Y}}) + \lambda \ell_{\text{REG}}.
\end{equation}
The following sections provide a comprehensive overview of these methods, which are summarized in Table \ref{table:noisylabelstructure}(b).

\subsubsection{Metric Learning Approaches}
Metric learning (ML) approaches treat the metric function as learnable parameters and refine the graph structures by learning the metric function $\mathcal \phi(\cdot, \cdot) $ of pair-wise representations: 
$\tilde{\mathbf{A}}_{ij} = \phi(\mathbf{h}_{i}, \mathbf{h}_{j}),$ 
% \begin{equation}
%     \tilde{\mathbf{A}}_{ij} = \phi(\mathbf{h}_{i}, \mathbf{h}_{j}),
% \end{equation}
where $\mathbf{h}_{i},\mathbf{h}_{j}$ are the learned embedding representations of nodes $v_i,v_j$, and $ \tilde{\mathbf{A}}_{ij}$ denotes the learned edge weight between $v_i$ and $v_j$. 
The refined matrix $\mathbf{A^*}$ is obtained as the output of an update function $ g( \cdot, \cdot) $: 
% \begin{equation}
    $\mathbf{A^*} = g(\mathbf{A}^{\text{noisy}}, \tilde{\mathbf{A}})$.
% \end{equation}
According to the different realizations of the metric function $\mathcal \phi(\cdot, \cdot) $, ML approaches can be divided into \emph{kernel-based} and \emph{neural-based MLs}. 
Additionally, $k$NN (i.e., each node has up to $k$ neighbors) and $\epsilon$NN (i.e., edges whose weight are less than $\epsilon$ will be removed) are two common post-processing operations to achieve graph sparsification.

\smallskip\textbf{Kernel-based ML (KML)}.
This kind of method uses the kernel function as $\phi$ to calculate edge weights between nodes. 
GRCN \cite{yu2021graph} contains a GCN that predicts missing edges and revises edge weights based on node embeddings. It uses the dot product as a kernel function to calculate the similarity between each node. Yet it cannot achieve reasonable graph sparsification. 
GNNGuard \cite{zhang2020gnnguard} protects GNNs from adversarial attacks by detecting and removing suspicious edges, ensuring robust predictions. It uses cosine similarity to assess connection relevance. 
Graph diffusion convolution (GDC) \cite{gasteiger2019diffusion} uses generalized graph diffusion for graph sparsification and improving learning outcomes, allowing for information aggregation from a broader neighborhood.
GDC uses a diffusion kernel function to quantify edge connections:
\begin{equation}
    \begin{aligned}
        \tilde{\mathbf{A}} =  \sum_{k=0}^{\infty}\theta_{k}\mathbf{T}^{k},
    \end{aligned}
\end{equation}
with the generalized transition matrix $\mathbf{T}$ and the weighting coefficients $\theta_{k}$ satisfying $\sum_{k=0}^{\infty}\theta_{k}=1$. 
Note that $\mathbf{T}$ can be the random walk transition matrix $\mathbf{T}_{\text{rw}}=\mathbf{A}^{\text{noisy}}\mathbf{D}^{-1}$ and the symmetric transition matrix $\mathbf{T}_{\text{sym}} = \mathbf{D}^{-1/2}\mathbf{A}^{\text{noisy}}\mathbf{D}^{-1/2}$, where $\mathbf{D}$ is the diagonal matrix of node degrees.

\smallskip\textbf{Neural-based ML (NML)}.
Compared to kernel-based approaches, neural-based approaches use more complex neural networks as the metric function $\mathcal \phi(\cdot, \cdot)$ to calculate edge weights between nodes and learn an optimized graph structure. 
GLCN \cite{jiang2019semi} seeks to improve the performance of GCN in semi-supervised learning tasks by learning an optimal graph structure. 
It utilizes a graph learning layer to calculate the similarity between two nodes and generates an optimal adaptive graph representation $\tilde{\mathbf{A}}$ for subsequent convolution operation. Formally, it learns a graph $\tilde{\mathbf{A}}$ as:
\begin{equation}
    \tilde{\mathbf{A}}_{ij} = \frac{\exp(\mathrm{ReLU}(\boldsymbol{\alpha}^{\top}\lvert \mathbf{z}_i-\mathbf{z}_j\rvert))}{\sum_{j=1}^{|\mathcal{V}|}\exp(\mathrm{ReLU}(\boldsymbol{\alpha}^{\top}\lvert \mathbf{z}_i-\mathbf{z}_j \rvert))},
\end{equation}
where $\boldsymbol{\alpha}$ is the learnable parameter vector. 
IDGL \cite{chen2020iterative} iteratively refines graph structure and GNN parameters to improve node embeddings and prediction accuracy. 
It uses weighted cosine similarity to optimize graph structure: 
% , similar to the multi-head attention mechanism.
\begin{equation}
        \tilde{\mathbf{A}}_{ij} = \frac{1}{m} \sum_{p=1}^{m}\tilde{\mathbf{A}}_{ij}^{p}, ~~
        \tilde{\mathbf{A}}_{ij}^{p} = \cos(\mathbf{w}_p \odot \mathbf{z}_i, \mathbf{w}_p \odot \mathbf{z}_j),
\end{equation}
where $ \odot $ denotes the hadamard product. Specifically, $\mathbf{w}$ is a learnable weight metric with $m$ perspectives, and IDGL calculates the weighted average of cosine similarity for each head.
SLAPS \cite{fatemi2021slaps} employs complex neural networks as a metric function to learn task-specific graph structures via self-supervision. Its graph generator infers structures with learnable parameters, while denoising autoencoders enhance supervision through feature denoising.

\subsubsection{Sampling-based Approaches}
\label{abilistic}
Sampling-based approaches involve randomly sampling edges or nodes from the original input graph according to a specific ability distribution to generate a refined graph structure, formulated as:
\begin{equation}
    \mathbf{A^*} = \mathcal{S}_{\text{SAM}}(\mathbf{A}^{\text{noisy}}, \mathbf{X}),
\end{equation}
where $\mathcal{S}_{\text{SAM}}(\cdot,\cdot)$ is designed based on the graph data itself or the specific task. 
This method allows for partial and random subset aggregation during GNN training, alleviating the structure noise and enhancing the model's robustness. 
Additionally, sampling approaches can be further categorized based on their relevance to downstream tasks.

\smallskip\textbf{Task-independent Approaches}.
This method involves sampling or dropping without considering their relation to downstream tasks. 
DropEdge \cite{rong2020dropedge} is an edge-level sampling technique that improves GCNs by randomly removing a certain portion of edges from the input graph during training. 
This technique acts as a form of unbiased data augmentation and reduces the intensity of message passing between nodes. However, However, the absence of targeted supervision limits its adaptability to specific downstream tasks.  
Instead, DropCONN \cite{chen2020enhancing} is a biased graph-sampling technique that aims to mitigate the effects of graph adversarial attacks.
It penalizes adversarial edge manipulations by constructing random and deformed subgraphs, introducing a significant regularization effect on graph learning. 

\smallskip\textbf{Task-dependent Approaches}.
% Instead of directly deforming the graph structure, task-dependent approaches seek feedback from downstream tasks to make improvements. 
% PTDNet \cite{luo2021learning} is an edge-level approach that aims to denoise graphs and enhance model generalization ability by dropping task-irrelevant edges through a parameterized network. 
% It also uses nuclear norm regularization to impose a low-rank constraint on the graph to ensure better generalization. 
% FastGCN \cite{chen2018fastgcn} is a node-level approach that improves the efficiency of training GCNs by sampling vertices according to their importance rather than uniformly. 
% It addresses the computational challenges associated with the recursive expansion of neighborhoods in GCNs without sacrificing accuracy. 
% Similarly, NeuralSparse \cite{zheng2020robust} selectively removes task-irrelevant edges using a deep parameterized neural network based on structural and non-structural information.
% GraphSAGE \cite{hamilton2017inductive}, one of the most famous GNN models, can also be considered as a neighbor-level sampling method. GraphSAGE generates node embeddings by sampling and aggregating features of their neighbors, which can generalize to unseen nodes and have an inductive performance.
Instead of directly deforming the graph structure, task-dependent approaches seek feedback from downstream tasks to make improvements. 
PTDNet~\cite{luo2021learning} is an edge-level method that denoises graphs and enhances generalization by dropping task-irrelevant edges via a parameterized network. 
It also employs nuclear norm regularization to enforce a low-rank constraint on the graph. 
FastGCN~\cite{chen2018fastgcn} is a node-level approach that improves GCN training efficiency by sampling vertices based on importance, mitigating recursive neighborhood expansion without sacrificing accuracy. 
Similarly, NeuralSparse~\cite{zheng2020robust} removes task-irrelevant edges using a deep neural network based on structural and non-structural information. 
% GraphSAGE~\cite{hamilton2017inductive}, a widely-used GNN model, can be viewed as a neighbor-level sampling method that generates node embeddings by sampling and aggregating features from neighbors, enabling inductive generalization to unseen nodes.

\subsubsection{Direct Optimization Approaches}
Direct optimization (DO) approaches consider the adjacency matrix learnable, which are optimized by applying specific regularization or optimization methods, including 
% \emph{sparsification},  
% \cite{wan2021graph, jin2020graph}, 
\emph{smoothness}, 
% \cite{jin2020graph, yang2019topology, gao2020exploring}; 
\emph{adversarial attack and defense}, and \emph{decoupling-based approaches}.   
% \cite{dong2021equivalence, he2022structural, chen2024pamt}, 

% \smallskip\textbf{Sparsification}. 
% This method is a typical regularization strategy that aims to reduce the complexity of a graph while retaining important information and minimizing the loss of related tasks.
% GSML \cite{wan2021graph} is a graph sparsification method that uses meta-learning principles for effective sparsification. 
% It treats graph structure as a hyper-parameter and selectively removes edges from the graph to maintain or even improve classification accuracy with a lighter graph structure. 

\smallskip\textbf{Smoothness}. 
This method is based on a commonly held assumption that graph signals change smoothly between adjacent nodes \cite{ortega2018graph}. 
This assumption typically refers to the smoothness of features and labels, assuming that nearby and connected nodes are likely to share the same labels or similar node features. 
% These methods rely on updating features and propagating labels as the core elements, which can also help to reduce the impact of structure noise.
TO-GCN \cite{yang2019topology} uses label smoothness regularization, jointly and alternately optimizing network topology and updating GCN parameters by fully utilizing the label and topology information. 
Pro-GNN \cite{jin2020graph} uses feature smoothness regularization to recover a clean graph structure, jointly updating GNN parameters assisted by other techniques like low rank and sparsification. 
% It aims to enhance the robustness of GNNs against adversarial attacks and improve node classification performance, thus denoising the graph structure.

\smallskip\textbf{Adversarial Attack and Defense}. 
This method handles structure noise by adding adversarial changes to graph edges \cite{sun2022adversarial}. 
Xu et al.~\cite{xu2019topology} introduces a gradient-based approach to help GNNs resist both greedy and gradient attacks without lowering accuracy. 
% Dwns~\cite{dai2019adversarial} takes an unsupervised approach. It learns perturbations by increasing the effect of random noise in the embedding space. 
Gosch et al.~\cite{gosch2023adversarial} leverages learnable graph diffusion to adaptively defend against structural perturbations while satisfying global and local constraints. 

% \smallskip\textbf{Community}. 
% This method assumes that nodes with similarities are more likely to be connected. 
% It is usually utilized to restrict the number of connected components in a graph, which is the essence of low-rank regularization.
% DHGR \cite{bi2022make} assumes that similar nodes should be connected by homophilic edges instead of heterophilic edges, which adds homophilic edges and prunes heterophilic edges based on node similarity in a label or feature distribution.
% % DHGR improves the performance of GNNs in heterophily graphs by rewiring the graph structure. This is done by adding homophilic edges and pruning heterophilic edges based on node similarity in a label or feature distribution.
% % DHGR adds homophilic edges and prunes heterophilic edges based on node similarity in a label or feature distribution.

\smallskip\textbf{Decoupling}. Decoupling-based methods decoupled the GNN architecture and refine the conventional cross-entropy loss. Decoupled GCN has been theoretically shown to be equivalent to label propagation~\cite{dong2021equivalence}, and is notably robust to structure noise. To overcome its sensitivity to initialization and label noise, PTA~\cite{dong2021equivalence} combines graph structural proximity and predictive confidence to dynamically reweight pseudo-labels in the cross-entropy loss, addressing both structure and label noise. 
To further integrate attribute information, RLP~\cite{he2022structural} adjusts the propagation matrix by combining attribute similarity with structural cues, and employs a momentum strategy for training stability. PAMT~\cite{chen2024pamt} uses a Hadamard product between an adaptive similarity mask and the adjacency matrix to mitigate inaccurate initial propagation caused by structure noise.

%% file: 6.3_attribute_noise.tex
\subsection{Attribute Noise}
\label{sec::attribute_noise}
Attribute noise refers to errors or disruptions in the features of nodes or edges in a graph. These perturbations can harm model performance by distorting the original feature distributions. Such noise often comes from data collection mistakes, missing values, or even intentional adversarial attacks~\cite{sun2022adversarial} that aim to corrupt the input features. The aim is to train a robust GNN $\mathcal{F}^*$ classifier that minimizes the impact of attribute perturbations $\delta$:
\begin{equation}
\mathcal{F}^* = \mathop{\arg\min}_{\mathcal{F}} ~\ell(\mathcal{F}(\mathbf{X}^{\text{noisy}}, \mathbf{A}), \mathbf{Y}). 
% \quad \|\delta\| \leq \epsilon,
\end{equation}
% where $\delta$ denotes bounded perturbations (e.g., $\ell_2$-norm constraints). 
Fundamental strategies against this noise fall into two categories: \emph{adversarial attack and defense} and \emph{loss refinement}. Table \ref{table:noisylabelstructure}(c) presents the overview of these methodologies.

\subsubsection{Adversarial Attack and Defense Approaches}
\label{subsec:adversarial_defense}
Adversarial attack and defense methods harden models against malicious perturbations through min-max optimization, formalized as:
\begin{equation}
\min_{\mathcal{F}} \max_{\|\delta\| \leq \epsilon} \ell(\mathcal{F}(\mathbf{X} + \delta, \mathbf{A}), \mathbf{Y}),
\end{equation}
where $\delta$ denotes adversarial noise constrained by $\epsilon$. For this type, several methods have been proposed to enhance the robustness of GNNs. For example, Nettack~\cite{zugner2018adversarial} is one of the first to reveal the vulnerability of GNNs. It generates subtle adversarial perturbations on both node features and graph structure. GraphAT~\cite{feng2019graph} introduces a dynamic regularizer. It helps stop perturbations from spreading too far in the graph. GCNVAT~\cite{sun2019virtual} looks at sensitive feature directions. It smooths GCNs along these directions to reduce the effect of adversarial changes. 
BVAT~\cite{deng2023batch} uses virtual adversarial training. It runs the training in batches to better learn local graph structures.  
GCORN~\cite{abbahaddou2024bounding} enhances inherent robustness against node feature attacks by enforcing orthonormal weight matrices, yielding an attack-agnostic robust GNN.

\subsubsection{Loss Refinement Approaches}
\label{subsec:loss_refinement}
Loss refinement techniques enhance robustness by modifying the objective function, typically through regularization terms grounded in stability analysis:
\begin{equation}
    \min_{\mathcal{F}} \ell_{\text{ERM}}(\mathcal{F}(\mathbf{X}^{\text{noisy}}, \mathbf{A}), \mathbf{Y}) + \lambda \ell_{\text{REG}},
\end{equation}
where $\ell_{\text{REG}}$ penalizes sensitivity to input perturbations (e.g., via Lipschitz constraints~\cite{gama2020stability}). T2-GNN~\cite{huo2023t2} proposes a dual teacher-student framework, where one teacher focuses on node features and the other on graph structure. Both teachers transfer clean patterns to assist in recovering corrupted attributes. MQE~\cite{li2024noise} takes a probabilistic view and learns noise-invariant meta-representations by estimating the quality of multi-hop features using a Gaussian model. Beyond explicit repair, BRGCL~\cite{wang2022bayesian} offers a different angle, embedding Bayesian nonparametrics into contrastive learning. The model iteratively distills robust prototypes from nodes with high confidence.

\subsection{Discussion}

% In addition to node-level robust GNN training against label noise, some studies target graph-level training, aiming to achieve reliable predictions for unlabeled graphs in the presence of noisy labels. For instance, D-GNN~\cite{nt2019learning} estimates MTM and applies backward loss correction to approximate clean labels. OMG~\cite{yin2023omg} combines coupled Mixup and graph CL, and introduces a neighbor-aware denoising strategy to enhance neighborhood smoothness. Despite these efforts, research on noisy labels in graph data remains limited, highlighting the need for novel strategies to improve model robustness. Furthermore, noisy labels may also arise in related tasks such as transfer learning~\cite{yuan2023alex}, imbalanced learning~\cite{pan2013graph}, and out-of-distribution detection~\cite{kim2023neural}, which deserve further exploration.

% D-GNN~\cite{nt2019learning} estimates MTM and applies backward loss correction to approximate clean labels. OMG~\cite{yin2023omg} combines coupled Mixup and graph CL, and introduces a neighbor-aware denoising strategy to enhance neighborhood smoothness.

Noise in real world GNNs mainly includes label noise, structure noise and attribute noise, all of which degrade model performance and robustness. Beyond node-level robust GNN training against label noise, some studies focus on graph-level learning, aiming to achieve reliable predictions for unlabeled graphs in the presence of noisy labels, such as D-GNN~\cite{nt2019learning} and OMG~\cite{yin2023omg}. Methods addressing structure noise include meta-learning \cite{wan2021graph}, adversarial attack \cite{dai2018adversarial, jin2020graph, zhang2020gnnguard}, graph revision \cite{yu2021graph,jiang2019semi}, graph sampling \cite{rong2020dropedge,chen2020enhancing}, and others \cite{ju2023glcc,zhao2023dynamic}. 
The main idea is to optimize the graph structure and enhance robustness. 
To understand model robustness (and inversely sensitivity), Lipschitz stability~\cite{gama2020stability} serves as a key metric for quantifying sensitivity to structure noise, influenced by both perturbation magnitude and Lipschitz constant. It provides theoretical guidance for designing graph models robust to structural noise. Such analysis can also guide the theoretical development of structure-resilient graph models. Despite these efforts, work that directly targets attribute noise is still scarce. Most existing research focuses more on label noise and structural noise, highlighting a critical gap in GNN literatures for real world.

%% file: 7_privacy.tex
\section{Privacy}

\begin{figure}
\centering
\includegraphics[width=\linewidth]{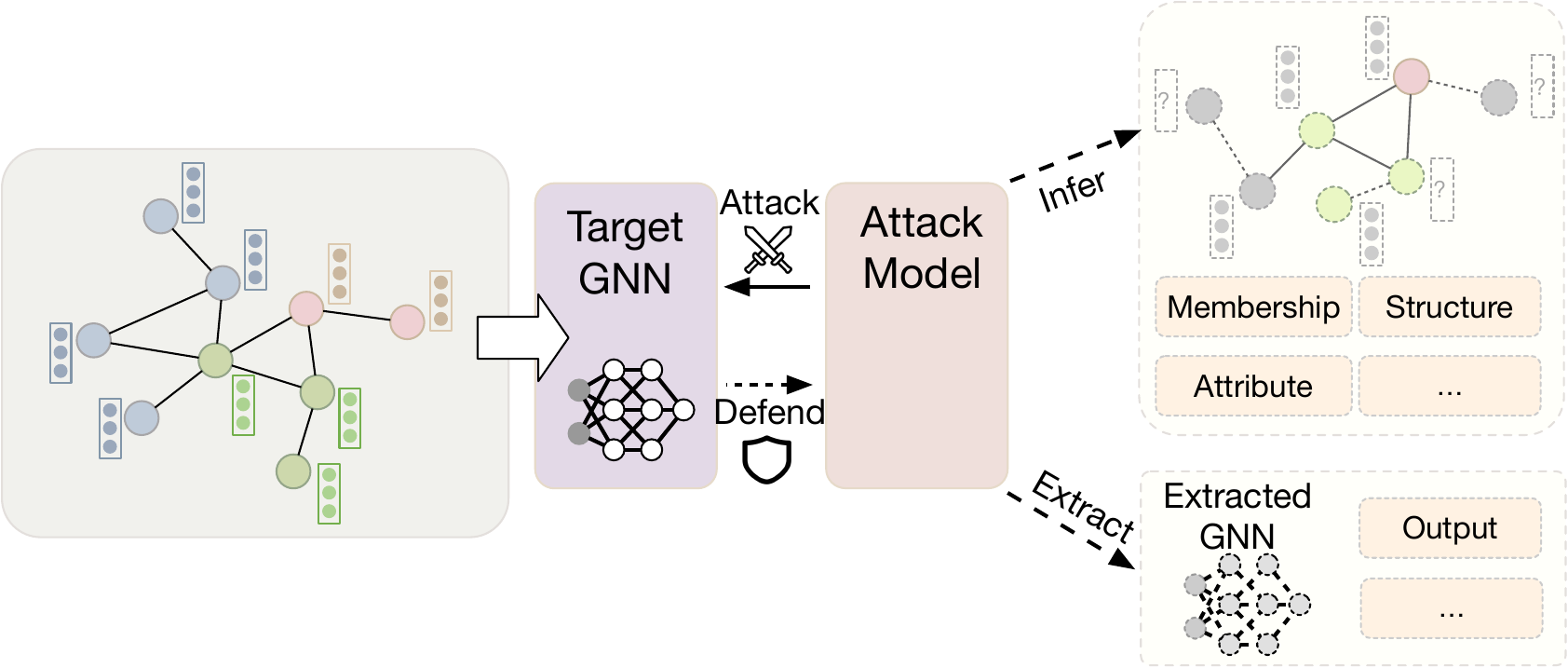}
\caption{Illustration of the attacks and defenses around both private data and model weights. The objective of the attack model is to extract private information from a target GNN. In response, the model needs to take measures and safeguard privacy from the attack model.}
\label{fig:privacy}
\end{figure}

% GNNs perform well on relational data and are widely used, but often underestimate privacy risks in sensitive fields like finance, e-commerce, social networks, and healthcare~\cite{zhang2024survey,dai2024comprehensive}. Like other deep models, most GNNs are vulnerable to privacy attacks~\cite{jagielski2020auditing}. Figure~\ref{fig:privacy} outlines a general GNN privacy framework, and Table~\ref{tab:privacy} summarizes key related works.
% %The following sections delve into each strategy, offering a comprehensive overview.
GNNs perform well on relational data and are widely used, but often overlook privacy risks in sensitive domains such as finance, e-commerce, and healthcare~\cite{zhang2024survey,dai2024comprehensive}. Like other deep models, most GNNs are vulnerable to privacy attacks~\cite{jagielski2020auditing}. Figure~\ref{fig:privacy} outlines a general GNN privacy framework, and Table~\ref{tab:privacy} summarizes related works.

\begin{table}[t]
\caption{Overview of methods for privacy attack/defend on GNNs models. (``SR'': structure reconstruction, ``AR'': attribute reconstruction, ``GPI'': general privacy issues)}
\label{tab:privacy}
\centering
\setlength{\tabcolsep}{0.1pt}
\resizebox{1\columnwidth}{!}{
\begin{tabular}{lccc}
\toprule
\multirow{2}{*}{Method} & Attack  & Focused  & Implementation \&  \\
& or Defend & Problem & Details \\
\midrule
%%%%%% Attack %%%%%%%%%
% Membership Inference
He et al.~\cite{he2021node} & Attack & MIA & Node-level, Black-box \\
He et al.~\cite{he2021stealing} & Attack & MIA & Edge-level, Black-box \\
Wu et al.~\cite{wu2021adapting} & Attack & MIA & Graph-level, Black-box \\
% Attribute Inference
Duddu et al.~\cite{duddu2020quantifying} & Attack & AIA & Node-level, Black-box \\
% Reconstruction
GraphMI~\cite{zhang2021graphmi} & Attack & RA & White-box, SR \\
% Model Extraction
Defazio et al.~\cite{defazio2019adversarial} & Attack & MEA & Adversarial Framework \\
Wu et al.~\cite{wu2022model} & Attack & MEA &  Information Leakage \\
\multirow{2}{*}{Shen et al.~\cite{shen2022model}} & \multirow{2}{*}{Attack} & \multirow{2}{*}{MEA} & Node-level, Black-box, \\
&&& Attack API access\\
%%%%%% Defend %%%%%%%%%
% DP
DPNE~\cite{xu2018dpne} & Defend & MIA & DP  \\
PrivGnn~\cite{olatunji2021releasing} & Defend & MIA & DP  \\
DP-GNN~\cite{mueller2022differentially} & Defend & MIA & DP  \\
KProp~\cite{sajadmanesh2021locally} & Defend & MIA & DP  \\
GERAI~\cite{zhang2021graph} & Defend & MIA & DP \\
% Latent Factor Disentangling
DP-GCN~\cite{hu2022learning} & Defend & AIA & LFD \\
DGCF~\cite{wang2020disentangled} & Defend & AIA & LFD \\
% Adversarial Training
GAL~\cite{liao2021information} & Defend & AR & AT \\
% Wang et al.~\cite{wang2021privacy} & Defend & Graph Structure Reconstruction & Adversarial Training \\
{APGE~\cite{li2020adversarial}} & Defend & AR & AT \& LFD \\ 
% Federated Learning
SpreadGNN~\cite{he2021spreadgnn} & Defend & GPI & FL  \\
D-FedGNN~\cite{pei2021decentralized} & Defend & GPI & FL  \\
%GCFL~\cite{xie2021federated} & Defend & General Privacy Issues & Federated Learning  \\
%GraphFL~\cite{wang2022graphfl} & Defend & General Privacy Issues & Federated Learning  \\
%GIB~\cite{wu2020graph} & Defend & General Privacy Issues & Federated Learning  \\
% FedPerGNN~\cite{wu2022federated} &  Defend & General Privacy Issues & Federated Learning  \\
% % Federated Learning + DP
% FedGNN~\cite{wu2021fedgnn} & Defend & Membership Inference & Federated Learning \& Differential Privacy \\
% VFGNN~\cite{chen2020vertically} & Defend & Membership Inference & Federated Learning \& Differential Privacy \\
% FeSoG~\cite{liu2022federated} & Defend & Membership Inference & Federated Learning \& Differential Privacy \\
% Machine Unlearning
GraphErase~\cite{chen2022graph} & Defend & GPI & Machine Unlearning \\
% GIF~\cite{wu2023gif} & Defend & General Privacy Issues & Machine Unlearning \\
% CEU~\cite{wu2023certified} & Defend & General Privacy Issues & Machine Unlearning \\
% Model Ownership Verification
watermark~\cite{xu2023watermarking,zhao2021watermarking} & Defend &  MEA & MOV \\
% Other
MIAGraph~\cite{olatunji2021membership} & Attack \& Defend & MIA &  DP \\
\bottomrule
\end{tabular}
}
\end{table}

\subsection{Privacy Attack}

% https://arxiv.org/pdf/2308.16375.pdf
% https://arxiv.org/pdf/2204.08570.pdf

Privacy attacks on GNNs target sensitive information -- such as training data, node/link attributes, or model parameters --and are typically classified into four categories based on their objectives~\cite{zhang2024survey,dai2024comprehensive}.

% membership inference attack (MIA)
\smallskip\textbf{Membership Inference Attack (MIA).}
MIAs aim to reveal whether a sample -- such as a node~\cite{he2021node}, edge~\cite{he2021stealing}, subgraph~\cite{olatunji2021membership}, or entire graph~\cite{wu2021adapting} -- is included in the training dataset, leading to potential information leakage~\cite{li2021membership}. 
Formally, with the GNN model $\mathcal{F}$ trained on training dataset $\mathcal{D}$, the attacker builds a binary classifier $\mathcal{A}_{\text{MIA}}$ such that:
\begin{equation}
    \mathcal{A}_{\text{MIA}}(\mathcal{F}(v)) = \begin{cases}
        1 & \text{if } v \in \mathcal{D}, \\
        0 & \text{otherwise}.
    \end{cases}
\end{equation}
For instance, MIAGraph~\cite{olatunji2021membership} trains a shadow model on data similar to the target model's training set and uses it to guide the attacker. The success of MIA largely relies on overfitting and information leakage from GNN, thus lacking consideration of the generalization gap.

% attribute/property inference attack (AIA)
% olatunji2023does
\smallskip\textbf{Attribute (/Property) Inference Attack (AIA).}
AIAs aim to uncover data attributes that are not explicitly included in the feature set~\cite{zhang2022inference,duddu2020quantifying}. 
Given a trained GNN model $\mathcal{F}$, a node $v_i \in \mathcal{D}$ with attributes $\mathbf{x}_i$, and its neighborhood $\mathcal{N}(v_i)$, 
% and given that the node $v$'s attributes are denoted as $x_v$. 
There are some public attributes $\mathbf{x}_i^\text{pub} \subset \mathbf{x}_i$ used as features, and some private attributes excluded from input features $\mathbf{x}_i^\text{priv} = \mathbf{x}_i \backslash \mathbf{x}_i^\text{pub}$. The attacker's goal is to build an inference model $\mathcal{A}_\text{AIA}$ such that
\begin{equation}
    \hat{\mathbf{x}}_i^\text{priv} = \mathcal{A}_\text{AIA}( \mathcal{F}, \mathbf{x}_i^\text{pub}, \{ \mathbf{x}_j \}_{v_j \in \mathcal{N}(v_i)} )
\end{equation}
can be used to estimate the private attributes of node $v_i$. If the attack is white-box, the attacker may also access the internal layers of $\mathcal{F}$, such as $\mathbf{h}_i^{(l)}$, a hidden representation of node $v_i$ at layer $l$, and take the form of $\hat{\mathbf{x}}_i^\text{priv} = \mathcal{A}_\text{AIA}( \mathbf{h}_i^{(l)} )$. 
When the graph structural information is well-protected and thus $\mathcal{N}(v_i)$ is unknown, attackers may learn to extract some global properties of the training graphs (e.g., average node degree, graph density, connectivity). 
% global properties $\psi({\mathcal{G}}_i)$ of the training graphs $\mathcal{D} = \{ \mathcal{G}_i \}_{i=1}^{|\mathcal{D}|}$ (e.g., average node degree, graph density, connectivity). 
% such as:
% \begin{equation}
%     \hat{\psi} = \mathcal{A}_\text{attr\_global}(\mathcal{M}).
% \end{equation}
Real-world datasets of all kinds can be targeted -- e.g., GNNs on molecules (revealing chemical bonds, atom types) or social networks (leaking private traits like gender, age)~\cite{duddu2020quantifying}.

% reconstruction attack/graph structure reconstruction (GSR)
\smallskip\textbf{Reconstruction (/Model Inversion) Attack (RA).}

RAs aim to infer private information of target samples and are generally categorized into attribute and structure reconstruction~\cite{dai2024comprehensive}. Attribute RAs recover node features $\mathbf{x}_i$ from embedding $\mathbf{h}_i$ via $\hat{\mathbf{x}}_i = \mathcal{A}_\text{ARA}(\mathbf{h}_i)$, while structure RAs infer graph topology, e.g., $\hat{\mathbf{A}}_{ij} = \mathcal{A}_\text{SRA}(\mathbf{h}_i^\top \mathbf{W} \mathbf{h}_j)$, where $\mathcal{A}_\text{SRA}$ can be a sigmoid function to get edge probabilities. Unlike AIAs, RAs target public attributes $\mathbf{x}_i^\text{pub}$ embedded in features~\cite{dai2024comprehensive}. These attacks often assume access to node embeddings~\cite{duddu2020quantifying,zhang2021graphmi,zhang2022model,wu2022linkteller}. For instance, GraphMI~\cite{zhang2021graphmi} employs projected gradient descent and a graph auto-encoder to reconstruct the adjacency matrix, and feature explanations can further enhance structure inference~\cite{olatunji2022private}.

% model extraction attack (MEA)
\smallskip\textbf{Model Extraction (/Stealing) Attack (MEA).} 
% MEAs pose a significant threat to large models accessed through APIs~\cite{niu2020dual}, potentially enabling other privacy and adversarial attacks~\cite{dai2024comprehensive}. Attackers aim to replicate the target model by learning a similar model that mimics its performance and decision boundaries.
% Let $\mathcal{Q} = \{ \mathcal{G}_i \}_{i=1}^{|\mathcal{Q}|}$ be a set of query graphs, and $\mathcal{F}(\mathcal{G}_i)$ be the predictions returned by the model, the attacker's goal is to train a surrogate model $\hat{\mathcal{F}}$ such that:
MEAs pose a significant threat to large models accessed via APIs~\cite{niu2020dual}, potentially enabling other privacy and adversarial attacks~\cite{dai2024comprehensive}. 
Attackers aim to replicate the target model by training a surrogate that mimics its performance and decision boundaries. 
Let $\mathcal{Q} = \{ \mathcal{G}_i \}_{i=1}^{|\mathcal{Q}|}$ be a set of query graphs, and $\mathcal{F}(\mathcal{G}_i)$ the predictions returned by the model. 
The attacker's goal is to train a surrogate model $\hat{\mathcal{F}}$ such that:
\begin{equation}
    \hat{\mathcal{F}}(\mathcal{G}_i) \approx \mathcal{F}(\mathcal{G}_i), ~~ \forall \mathcal{G}_i \in \mathcal{Q}.
\end{equation}
Some privacy attacks operate in a transductive or white-box setting, while others follow an inductive or black-box paradigm~\cite{rong2020self}. Early methods using adversarial frameworks achieved up to 80\% output similarity~\cite{defazio2019adversarial}, whereas more recent approaches report fidelity as high as 90\% on transductive GNNs~\cite{wu2022model}. MEAs on inductive GNNs have also shown strong effectiveness, even when attackers are restricted to remote API access to the victim models~\cite{shen2022model}.

\subsection{Privacy Preservation}

On the other side, various methods have been proposed to make GNNs more resilient to privacy attacks~\cite{zhang2024survey}.

\smallskip\textbf{Differential Privacy (DP).}
DP offers formal privacy guarantees for both i.i.d. and graph data. It ensures that an algorithm's output remains nearly unchanged when applied to neighboring datasets $\mathcal{D}'$ differing from original dataset $\mathcal{D}$ by only a few records.
Formally, a randomized algorithm $\mathcal{M}$ satisfies $(\varepsilon, \delta)$-DF, if for all measurable subsets $S$ of outputs: 
\begin{equation}
    \mathbb{P}[\mathcal{M}(\mathcal{D}) \in S] \leq e^\varepsilon \cdot\mathbb{P}[\mathcal{M}(\mathcal{D}') \in S] + \delta.
\end{equation}
DP mitigates MIA by limiting individual data influence through noise injection, with strong theoretical guarantees~\cite{dwork2006calibrating}. 
% DPNE~\cite{xu2018dpne} applies objective perturbation to matrix factorization, aligning with embedding methods like DeepWalk and LINE. 
DPNE~\cite{xu2018dpne} enforces DP in network embedding via perturbed matrix factorization that implicitly preserves DeepWalk/LINE properties, though suffers significant utility loss ($\sim$30\% accuracy drop at $\epsilon=1$) due to high gradient sensitivity in random walks.
MIAGraph~\cite{olatunji2021membership} combines output perturbation and homophily reduction. 
PrivGNN~\cite{olatunji2021releasing} trains a DP-protected teacher model on poisoned data, then distills it into a student model. 
DP-GNN~\cite{mueller2022differentially} uses DP-SGD to privatize gradient updates. 
KProp~\cite{sajadmanesh2021locally} adds noise to node features pre-aggregation, relying on averaging to maintain utility. 
GERAI~\cite{zhang2021graph} addresses membership privacy in recommendation systems via dual-stage encryption, perturbing user features while optimizing a modified loss.

\smallskip\textbf{Latent Factor Disentangling (LFD).}
In typical GNNs, embeddings encode both sensitive and task-relevant information~\cite{li2020adversarial}. 
LFD addresses this by decomposing node embedding $\mathbf{h}_i$ into private and task-related components: 
$\mathbf{h}_i = [\mathbf{h}_i^\text{priv} || \mathbf{h}_i^\text{pub}],$ 
where $\mathbf{h}_i^\text{priv}$ captures private (sensitive) features, and $\mathbf{h}_i^\text{pub}$ captures utility (non-sensitive or public) features. Disentanglement enforces independence between them by minimizing mutual information, exactly corresponding to processing AIA. 
% APGE~\cite{li2020adversarial}, built on a graph autoencoder, augments the decoder with privacy-related labels to encourage label-invariant embeddings after disentangling. 
% In node-dependent privacy settings, DP-GCN~\cite{hu2022learning} introduces a two-module framework: one disentangles sensitive and non-sensitive representations (enforcing orthogonality), and the other trains a GCN on non-sensitive components for downstream tasks. It enhances privacy for users with hidden attributes (e.g., undisclosed age) by leveraging public users’ data. 
% DGCF~\cite{wang2020disentangled}, originally for graph collaborative filtering, also yields privacy-preserving embeddings by separating user intentions.
APGE~\cite{li2020adversarial}, built on a graph autoencoder (GAE), augments the decoder with privacy-related labels to encourage label-invariant embeddings. 
In node-dependent privacy settings, DP-GCN~\cite{hu2022learning} introduces a two-module framework: one disentangles sensitive and non-sensitive representations, and the other trains a GCN on non-sensitive components for downstream tasks. 
% It enhances privacy for users with hidden attributes (e.g., undisclosed age) by leveraging public data. 
DGCF~\cite{wang2020disentangled}, originally for graph collaborative filtering, also yields privacy-preserving embeddings by separating user intentions.

\smallskip\textbf{Adversarial Training (AT).}
A common defense approach is to deliberately reduce the effectiveness of specific privacy attacks. This strategy is known as AT~\cite{zhang2024survey} or adversarial privacy-preserving~\cite{dai2024comprehensive}.
This approach trains models to minimize attack success while preserving downstream task performance, typically framed as a min-max optimization where a GNN encoder $\mathcal{F}$ learns embeddings that are both task-relevant and privacy-preserving:
\begin{equation}
    \min_{\mathcal{F}} \max_{\mathcal{S}} \ell_\text{ERM} (\mathcal{F}(\mathcal{G}), \mathbf{Y}) - \lambda \ell_\text{ADV} (\mathcal{S}(\mathcal{F}(\mathcal{G})), \mathbf{Y}),
\end{equation}
% where $\ell_\text{ERM}$ is the empirical risk for a downstream task (e.g., node classification), $\mathcal{S}_\phi$ is an adversary model parameterized by $\phi$ attempting to infer private information from the learned embeddings, $\ell_\text{ADV}$ is the adversary's loss (e.g., for predicting sensitive attributes), $\lambda$ is a hyper-parameter that controls the trade-off between utility and privacy. 
% GAL~\cite{liao2021information} defends against worst-case attackers via graph information obfuscation. 
% APGE~\cite{li2020adversarial}, though framed as a disentangling method, also employs adversarial training through an adversarial autoencoder to learn privacy-preserving embeddings. 
% NetFense~\cite{hsieh2021netfense} instead perturbs the graph structure (e.g., the adjacency matrix) to mislead attackers without adversarially updating the model.
where $\ell_\text{ERM}$ is the empirical risk for a downstream task, $\mathcal{S}$ is an adversary model to infer private information from embeddings, $\ell_{\text{ADV}}$ is the adversary's loss (e.g., predicting sensitive attributes), and $\lambda$ is a hyper-parameter controlling the utility-privacy trade-off. 
GAL~\cite{liao2021information} defends against worst-case attackers via graph information obfuscation.
APGE~\cite{li2020adversarial}, framed as a LFD method, also uses AT via an adversarial autoencoder to learn privacy-preserving embeddings.
NetFense~\cite{hsieh2021netfense} instead perturbs the graph structure (e.g., the adjacency matrix) to mislead attackers without adversarial model updates. 
Thus, AT is capable of addressing various adversarial modes, such as AIA, RA, MEA, etc.
%From the perspective of making only the perturbed data available to the attackers, NetFense adopted the spirit of DP.

\smallskip\textbf{Federated Learning (FL).}
FL enables collaborative model training across distributed clients without sharing raw data, preserving privacy by aggregating local updates for global model optimization~\cite{kairouz2021advances, mcmahan2017communication}. 
Mathematically, let $\mathcal{K}$ be the set of participating clients, $\mathcal{D}$ be the full dataset, $\mathbf{w}$ be the global model parameters, $\mathbf{w}_k$ be the local model parameters on client $k \in \mathcal{K}$, $\ell_k(\mathbf{w}_k)$ be the loss on client $k$ over their private dataset $\mathcal{D}_k$, the federated objective is typically:
\begin{equation}
    \min_{\mathbf{w}} \sum_{k \in \mathcal{K}} \frac{|\mathcal{D}_k|}{|\mathcal{D}|} \cdot \ell_k(\mathbf{w}),
\end{equation}
where in each training round $t$, the federated averaging (FedAvg) algorithm proceeds as:
\begin{equation}
    \begin{aligned}
        \mathbf{w}_k^{(t+1)} & \leftarrow \mathbf{w}^{(t)} - \eta \nabla \ell_k(\mathbf{w}^{(t)}) , \qquad\text{(local update)} \\
        \mathbf{w}^{(t+1)} & \leftarrow \sum_{k \in \mathcal{K}} \frac{|\mathcal{D}_k|}{|\mathcal{D}|} \cdot \mathbf{w}_k^{(t+1)}.  \qquad\text{(global aggregation)} 
    \end{aligned}
\end{equation}
% This framework ensures that only the local server accesses raw data, preventing potential information leakage. Graph FL models are typically categorized into three types based on graph data distribution: (i) Inter-graph FL, where each client holds a subset of graph samples~\cite{he2021fedgraphnn}, (ii) Intra-graph FL, where each client owns a subgraph~\cite{yao2023fedgcn}, (iii) Decentralized FL, where clients communicate directly and aggregate information without a central server~\cite{zhang2021federated}. 
% FL faces challenges, such as training with partial labels in decentralized settings. SpreadGNN~\cite{he2021spreadgnn} addresses this with DPA-SGD, while D-FedGNN~\cite{pei2021decentralized} uses DP-SGD for a similar solution.
This framework ensures that only the local server accesses raw data, preventing potential information leakage. So it performs well in dealing with MIA and even more general privacy issues. 
Graph FL models are typically categorized into three types based on graph data distribution: 
(i) Inter-graph FL, where each client holds a subset of graph samples~\cite{he2021fedgraphnn}; 
(ii) Intra-graph FL, where each client owns a subgraph~\cite{yao2023fedgcn}; 
(iii) Decentralized FL, where clients communicate directly and aggregate without a central server~\cite{zhang2021federated}. 
FL faces challenges such as training with partial labels in decentralized settings. 
SpreadGNN~\cite{he2021spreadgnn} addresses this with DPA-SGD, while D-FedGNN~\cite{pei2021decentralized} adopts DP-SGD for a similar solution.

\subsection{Discussion}

% In addition to the methods discussed, other privacy-preserving techniques have been developed, such as machine unlearning~\cite{chen2022graph}, with GraphErase~\cite{chen2022graph} using a balanced graph partition to maintain performance after node removal. Another approach, model-ownership verification~\cite{xu2023watermarking,zhao2021watermarking}, embeds watermarks into models to protect intellectual property. As privacy concerns grow, especially with proprietary models or those trained on sensitive data, understanding and defending against privacy attacks will be crucial.

% Several defense strategies have been proposed for GNNs, each with trade-offs. Differential Privacy adds noise for strong guarantees but may reduce utility. Latent factor disentangling separates sensitive and task-relevant features for better control, though it requires prior knowledge. Adversarial training simulates attacks to balance defense and accuracy but can be unstable. Federated Learning avoids data sharing to preserve privacy but remains vulnerable without enhancements like DP. These methods complement each other depending on the threat model and use case.
In addition to the methods discussed, other privacy-preserving techniques have been developed, such as machine unlearning~\cite{chen2022graph}, where GraphErase~\cite{chen2022graph} uses balanced graph partitioning to maintain performance after node removal. 
Model-ownership verification (MOV)~\cite{xu2023watermarking,zhao2021watermarking} embeds watermarks into models to protect intellectual property. As privacy concerns grow, especially for proprietary models or those trained on sensitive data, understanding and defending against privacy attacks is crucial. Several defense strategies for GNNs exist, each with trade-offs: DP adds noise for strong guarantees but may reduce utility; LFD separates sensitive from task-relevant features for better control but requires prior knowledge; AT simulates attacks to balance defense and accuracy but can be unstable; FL avoids data sharing to preserve privacy but remains vulnerable without enhancements like DP. These methods complement each other depending on threat models and use cases.

%% file: 8.1_ood_detection.tex
\section{Out-of-distribution}

\begin{figure}
\centering
\includegraphics[width=\linewidth]{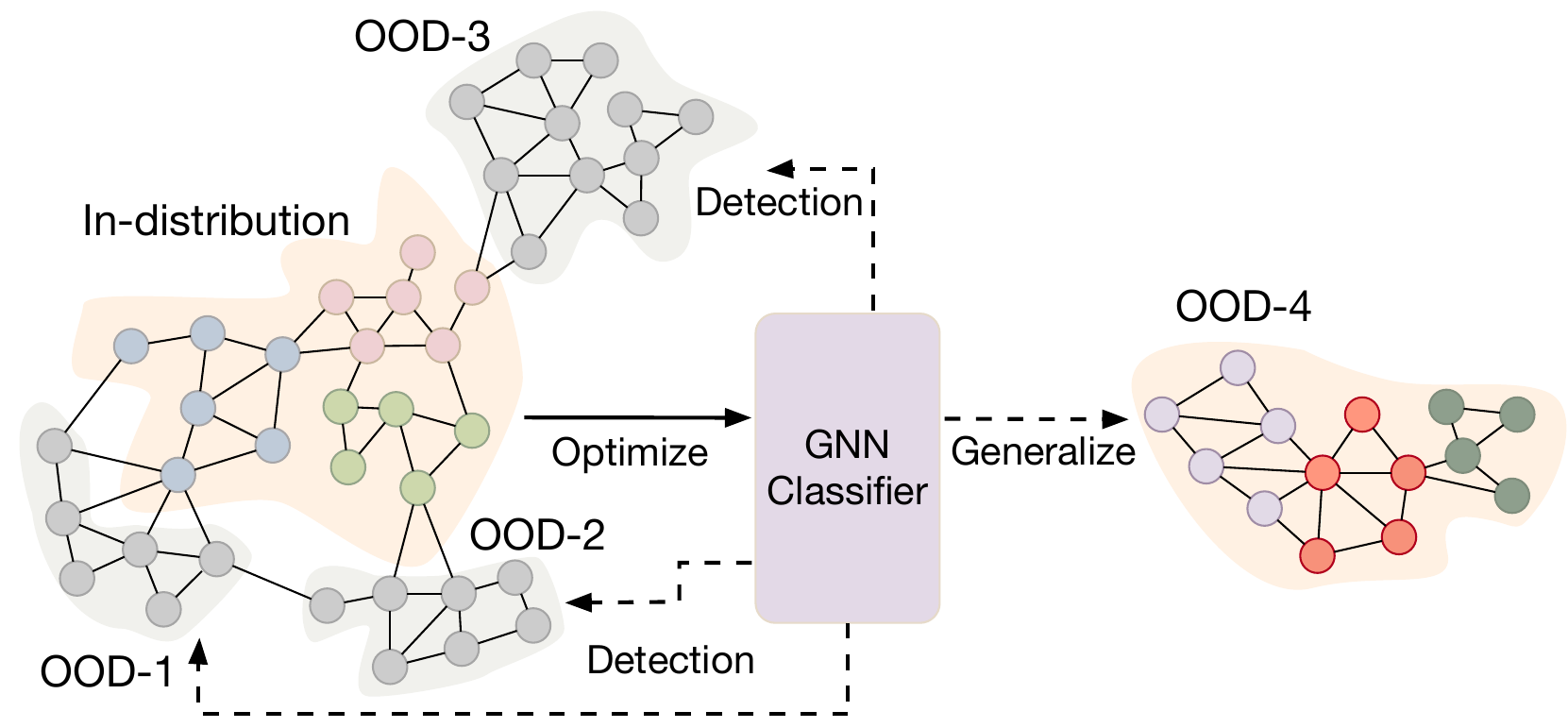}
\caption{Illustration of the OOD issue in real-world GNN training. While the model is trained on observed in-distribution data, the presence of OOD data calls for the development of mechanisms for OOD detection and generalization.}
\label{fig:OOD}
\end{figure}

Despite the strong representation capabilities of GNNs, they often display a mix of unsuitability and overconfidence when test sample distribution significantly deviates from the distribution of training samples. In this section, we delve into the Out-Of-Distribution (OOD) problem on graphs. There are two common OOD scenarios: \emph{OOD detection} and \emph{OOD generalization}. Figure \ref{fig:OOD} illustrates the basic schematic diagrams for these two scenarios in GNNs. Next, we will discuss these two aspects in detail.

\subsection{Out-of-distribution Detection}
OOD detection on graphs, which aims to distinguish test samples from the major in-distribution (ID) training data, has become an essential problem in real-world applications. Formally, we assume there is an ID graph dataset $\mathcal{D}^{\text{in}}=\{D_1^{\text{in}}, \cdots\, D_{N_1}^{\text{in}}\}$ and an OOD graph dataset $\mathcal{D}^{\text{out}}=\{D_1^{\text{out}}, \cdots\, D_{N_2}^{\text{out}}\}$, where data are sampled from a major distribution $\mathbb{P}^{\text{in}}$ and an OOD distribution $\mathbb{P}^{\text{out}}$, respectively. The general purpose of OOD detection on graphs is to identify its source distribution (i.e., $\mathbb{P}^{\text{in}}$ or $\mathbb{P}^{\text{out}}$) based on the learned detector $\mathcal{T}$: 
\begin{equation}
\mathcal{T}(D_v;\tau,s,\mathcal{F})=
\begin{cases}
    D_v\in\mathbb{P}^{\text{in}}, &\text{if}\quad s(D_v,\mathcal{F})\leq\tau\\
    D_v\in\mathbb{P}^{\text{out}}, &\text{if}\quad s(D_v,\mathcal{F})>\tau
\end{cases},
\end{equation}
where $\mathcal{F}$ is the trained model, $s$ is a scoring function and $\tau$ is the corresponding threshold. $D_v$ can be a node or a graph corresponding to the node-level or the graph-level task.

Based on the different scoring function designs, existing OOD detection methods on graphs can be roughly partitioned into: \emph{propagation-based} approaches, \emph{classification-based} approaches and \emph{self-supervised learning-based} approaches. Table \ref{tab:ood detection generalization}(a) presents the overview of these methods and we present a comprehensive introduction of these as follows.

\subsubsection{Propagation-based Approaches}
% Different from OOD detection in computation vision and natural language processing where OOD samples are i.i.d. and typically only appear in the test set, ID and OOD nodes are connected in one graph for node-level OOD detection. The core motivation of propagation-based approaches is to borrow the idea of label propagation (LP) or message propagation in GNN to transfer some of the existing uncertainty estimation (UE) definitions. 
Unlike i.i.d. OOD detection in CV/NLP, node-level OOD detection involves connected ID and OOD nodes. Propagation-based methods adapt label propagation or GNN message passing to transfer uncertainty estimation (UE) frameworks. The process is typically represented as:
\begin{equation}
    s_i^{(t)}=\alpha s_i^{(t-1)}+(1-\alpha)\sum_{v_j \in\mathcal{N}(v_i)}\Pi_{i,j} s_j^{(t-1)},
\end{equation}
where $\Pi_{i,j}$ reflects the importance of neighbor node $v_j$ on $v_i$, $\alpha$ controls the concentration parameter of scoring. GPN~\cite{stadler2021graph} explores uncertainty quantification for OOD node detection. The method extends the input-dependent Bayesian update and explicitly models epistemic and aleatoric uncertainty by propagating node-wise estimates along the graph. However, Bayesian-based methods typically suffer from high computational complexity, which has motivated research into alternative approaches and approximate implementations. 
GNNSage~\cite{wu2023energy} proposes a node-level OOD detection based on an energy function and introduces an energy-based belief propagation (BeP), which propagates the estimated energy score among nodes in the graph iteratively. 
OODGAT~\cite{song2022learning} explicitly models the interaction between ID and OOD nodes and separates these two types of nodes during feature propagation. 
OSSNC~\cite{huang2022end} learns to mix neighbors to mitigate the propagation to and from OOD nodes in a variational inference (VI) framework for simultaneous node classification and OOD detection.

\subsubsection{Classification-based Approaches}
Another typical OOD detection method originated from a simple baseline, which uses the maximum softmax probability as the indicator scores of ID-ness~\cite{hendrycks2017baseline}. The formation of the OOD score can be:
\begin{equation}
    s(D_v,\mathcal{F}) = \operatorname{Max}(\mathcal{F}(D_v)).
\end{equation}
AAGOD~\cite{guo2023data} proposes a data-centric post-hoc method instead of re-training the model for graph-level OOD detection. The method adopts a learnable amplifier generator to enlarge the indicator score gap between OOD and ID graphs. 
Some classification-based approaches also focus on node-level OOD detection and graph anomaly detection. 
BWGNN~\cite{tang2022rethinking} uses the Beta wavelet kernel as a tailored spectral filter in GNN for node anomaly detection. GKDE~\cite{zhao2020uncertainty} considers multidimensional uncertainties for node-level OOD detection. iGAD~\cite{zhang2022dual} treats graph-level anomaly detection as a special case of graph classification and proposes a dual-discriminative framework with GNN and graph kernel together to learn the label.

\begin{table}[t]
\caption{Overview of methods for graph OOD detection and generalization. 
(``BaP'': Bayesian posterior, ``BLO'': bi-level optimization, ``SI'': statistical independence, ``GNAS'': graph neural architecture search)}
\label{tab:ood detection generalization}
\centering

{\bf (a).} OOD Detection
\vspace{2mm}

\setlength{\tabcolsep}{3pt}
\resizebox{1\columnwidth}{!}{
\begin{tabular}{lccc}
\toprule
Method & Task Type & Core Idea & Implementation \& Details \\
\midrule
GPN~\cite{stadler2021graph} & Node-level & Propagation & BaP, UE \\
GNNSage~\cite{wu2023energy}  & Node-level & Propagation & Energy, BeP \\
OODGAT~\cite{song2022learning} & Node-level & Propagation & Entropy Regularization\\
OSSNC~\cite{huang2022end} & Node-level & Propagation & VI, BLO \\
AAGOD~\cite{guo2023data} & Graph-level & Classification & Data-Centric, Amplifier \\
BWGNN~\cite{tang2022rethinking} & Node-level & Classification & Graph
Wavelet \\
GKDE~\cite{zhao2020uncertainty} & Node-level & Classification & UE \\
iGAD~\cite{zhang2022dual} & Node-level & Classification & Graph Kernel \\
GLocalKD~\cite{ma2022deep} & Graph-level & SSL & KD \\
GOOD-D~\cite{liu2023good} & Graph-level & SSL & CL \\
GRADATE~\cite{duan2023graph} & Graph-level & SSL & CL \\
GLADC~\cite{luo2022deep} & Graph-level &  SSL & GAE, GR \\
GraphDE~\cite{li2022graphde} & Node-level &  SSL & VI, GR\\
OCGIN~\cite{zhao2023using} & Graph-level & SSL & OCC \\
OCGTL~\cite{qiu2023raising} & Graph-level & SSL & OCC \\
GOODAT~\cite{liu2023good} & Graph-level & SSL & IB \\
SIGNET~\cite{liu2023towards} & Graph-level & SSL & IB, Hypergraph \\
SGOOD~\cite{ding2023sgood} & Graph-level & SSL & Substructure \\
\bottomrule
\end{tabular}
}

\vspace{2mm}
{\bf (b).} OOD Generalization
\vspace{2mm}

\setlength{\tabcolsep}{2pt}
\resizebox{1\columnwidth}{!}{
\begin{tabular}{lccc}
\toprule
Method & Task Type & Core Idea & Implementation \& Details \\
\midrule
DIR \cite{wu2022discovering} & Graph-level & Subgraph & CI, Intervention \\
CAL~\cite{sui2022causal}  & Graph-level & Subgraph & CI, Disentanglement \\
CIGA~\cite{chen2022learning} & Graph-level & Subgraph & CI, CL \\
StableGNN~\cite{fan2023generalizing} & Graph-level & Subgraph & CI, Regularization \\
SizeShiftReg~\cite{buffelli2022sizeshiftreg} & Graph-level & Subgraph & IL, Regularization \\
GIL~\cite{li2022learning} & Graph-level & Subgraph & IL, Regularization \\
FLOOD~\cite{liu2023flood} & Node-level & Subgraph & IL, CL \\
LiSA~\cite{yu2023mind} & Graph \& Node  & Subgraph & IL, Regularization \\
EERM~\cite{wu2022handling} & Node-level & Subgraph & IL, AL \\
GraphAT~\cite{feng2019graph} & Node-level & AL & Augmentation \\
CAP~\cite{xue2021cap} & Node-level & AL & Augmentation \\
\multirow{2}{*}{WT-AWP~\cite{wu2023adversarial}} & \multirow{2}{*}{Node-level} & \multirow{2}{*}{AL} & Augmentation, \\
&&& Regularization \\
LECI~\cite{gui2023joint} & Graph-level & AL & IL, Augmentation \\
AIA~\cite{sui2023unleashing} & Graph-level & AL & Augmentation \\
OOD-GNN~\cite{li2022ood} & Graph-level &  RD & SI \\
GRACES~\cite{qin2022graph} & Graph-level &  RD & GNAS \\
OOD-LP~\cite{zhou2022ood} & Edge-level &  TA & - \\
\bottomrule
\end{tabular}
}
\end{table}

\subsubsection{Self-supervised Learning-based Approaches}
Since data labeling on graph-structured data is commonly time-consuming and labor-intensive~\cite{ju2022kgnn}, recent studies also consider the scarcity of class labels and OOD samples. The basic idea is to learn a self-supervised learning (SSL) framework for OOD detection on graphs based on unlabeled ID data and the method is mainly focused on graph-level OOD detection or anomaly detection. 

\smallskip\textbf{Contrastive Learning (CL).} A popular approach for self-supervised graph OOD detection is to drive multiple views of a graph sample and detect the OOD sample based on inconsistency.   
GLocalKD~\cite{ma2022deep} jointly learns two GNNs and performs graph-level and node-level random knowledge distillation (KD) between the learned representations of two GNNS to learn a graph-level anomaly detector. But the singularity of perspective and scale limits its semantic richness.
GOOD-D~\cite{liu2023good} performs perturbation-free graph data augmentation and utilizes hierarchical CL on the generated graphs for graph-level OOD detection. 
GRADATE~\cite{duan2023graph} proposes a multi-view multi-scale CL framework with node-node, node-subgraph and subgraph-subgraph contrast for graph anomaly detection.

\smallskip\textbf{Graph Reconstruction (GR).} Some works also aim to discriminative representations through reconstruction mechanisms and infer the graph OOD samples. GLADC~\cite{luo2022deep} uses graph CL to learn node-level and graph-level representations and measure anomalous graphs with the error between generated reconstruction graph representations and original graph representations in a graph convolution autoencoder way. GraphDE~\cite{li2022graphde} models the generative process of the graph to characterize the distribution shifts. Thus, ID and OOD graphs from different distributions indicate different environments and can be inferred by VI. 

\smallskip\textbf{One-Class Classification (OCC).} The goal of OCC is to train embeddings to cluster within a defined hypersphere, establishing a decision boundary. OCGIN~\cite{zhao2023using} studies an end-to-end GNN model with OCC for anomaly detection. OCGTL~\cite{qiu2023raising} further extends the deep OCC method to a self-supervised detection way using neural transformations graph transformation learning as regularization. GOODAT~\cite{wang2024goodat} introduces a graph test-time OOD detection method under the graph information bottleneck (IB) principle to capture informative subgraphs. The surrogate labels are inherently ID, which can be seen as another kind of OCC.

% \subsubsection{Discussion}
% Researches of OOD detection on graphs explore different scoring functions to identify the OOD data. In addition to the above methods, some works also extend OOD detection to be more explainable. Especially graph-level OOD detection and anomaly detection, the methods provide meaningful explanations for the detection scores. For example, SIGNET~\cite{liu2023towards} proposes a self-interpretable graph-level anomaly detection framework that infers graph-level anomaly scores and provides the subgraph explanation simultaneously via maximizing the mutual information of the constructed multi-view subgraphs. SGOOD~\cite{ding2023sgood} explicitly utilizes graph substructures and their relationships to learn substructure-enhanced graph representations for graph-level OOD detection. 

%% file: 8.2_ood_generalization.tex
\subsection{Out-of-distribution Generalization}

Another key challenge in real-world OOD scenarios is graph OOD generalization~\cite{gui2022good,li2022out}, addressing distribution shifts between training/test data. It covers node-level (node classification) and graph-level tasks (graph classification), with the latter being more studied. Let $\mathcal{F}$ denote the GNN classifier.The objective is to find the optimal $\mathcal{F}^*$ satisfying: 
\begin{equation*}
\mathcal{F}^{*}=\underset{\mathcal{F}}{\arg \min } \sup _{e \in \mathit{E}} \mathbb{E}_{(\mathcal{G},y)\in \mathcal{S}^e}[\ell(\mathcal{F}(\mathcal{G}), y)],
\end{equation*}
where $\mathit{E}$ is a set to collect all the test environments, and $\mathcal{S}^e$ include all graph-label pairs in the environment $e$. $\ell(\mathcal{F}(\mathcal{G}), y)$ calculates the tailored loss for each sample. 
Distribution shift typically involves attributive shift (node attribute changes from varying backgrounds/environments) and structural shift (adjacency matrix variations from connectivity/graph size differences).  Table \ref{tab:ood detection generalization}(b) presents a overview of these methods and we introduce these below.

\subsubsection{Subgraph-based Approaches}

Subgraph-based approaches~\cite{wu2022discovering} assume that every graph consists of a crucial part and a non-crucial part from semantic and environmental information, respectively. To identify subgraphs with crucial knowledge, they usually utilize causal inference and invariant theory for effective graph representation learning.

\smallskip\textbf{Causal Inference (CI).} A key research direction constructs a structural causal graph (SCG) for theoretical analysis (TA), modeling interactions between invariant and spurious components. Following \cite{wu2022discovering}, invariant component is typically extracted via learnable subgraph masking, with loss functions enforcing invariance to non-causal features, which can result in a common form of the loss objectives as:
\begin{equation}\label{eq:intervener}
    \min \ell_{\text{ERM}}(\mathcal{F}(\mathcal{G}),\mathbf{Y}) + \lambda\ell_{\text{VAR}},
\end{equation}
where $\ell_{\text{ERM}}$ denotes the empirical risk on the training dataset and $\ell_{\text{VAR}}$ is related to the variances of the predictions with different simulated spurious factors viewed as intervention. Based on this framework, numerous advanced variants are developed by integrating different techniques. 
For example, CAL~\cite{sui2022causal} incorporates graph representations into the SCG, followed by the attention mechanism and representation disentanglement to select causal patterns. Besides, shortcut features are included in graph representations using the backdoor adjustment theory. CAL's attention-based causal selection lacks theoretical invariance guarantees and relies on predefined causal assumptions.  
Further, CIGA~\cite{chen2022learning} considers the graph generation process with and without partial interaction between invariant and spurious parts, and then identifies the crucial subgraph by maximizing the intra-class semantics for invariance. 
In addition, StableGNN~\cite{fan2023generalizing} adopts a differentiable graph pooling operator for subgraph extraction, which is optimized using a distinguishing regularizer with reduced spurious correlations.

% \smallskip\textbf{Invariant Theory (IT).}  
% As in Eq.~\eqref{eq:intervener}, causality-based approaches connect IT with intervention. More other methods use IT through data augmentation to create diverse training environments, extending empirical risk minimization to invariant risk minimization for distribution shift robustness. 
% A general idea is to extend empirical risk minimization into invariant risk minimization to increase the robustness to the distribution shift. 
% For example, SizeShiftReg~\cite{buffelli2022sizeshiftreg} simulates the size shift using graph coarsening and proposes a simple regularization loss for consistency learning after coarsening. GIL~\cite{li2022learning} learns a mask matrix for subgraph generation and enforces model invariance to environment inference via an invariance regularizer. FLOOD~\cite{liu2023flood} generates augmented graphs using node dropping and attribute masking and then conducts bootstrapped learning with a CL architecture~\cite{grill2020bootstrap}. 
% MoleOOD~\cite{yang2022learning} produces environment variance using VI, which guides the invariant learning (IL) to improve the OOD generalization capacity. 
% LiSA~\cite{yu2023mind} builds variational subgraph generators with information constraints to simulate diverse environments and promotes subgraph diversity via energy-based regularization. 
% EERM~\cite{wu2022handling} introduces different contexts to simulate virtual environments, which are trained adversarially for node-level IL. 
\smallskip\textbf{Invariant Theory (IT).}  
As in Eq.~\eqref{eq:intervener}, causality-based methods relate IT to interventions. Other approaches leverage IT via data augmentation to improve robustness against distribution shifts by extending ERM to invariant risk minimization. For instance, SizeShiftReg~\cite{buffelli2022sizeshiftreg} simulates size shifts with graph coarsening and applies a regularization loss for consistency. GIL~\cite{li2022learning} learns subgraph masks and uses invariance regularization. FLOOD~\cite{liu2023flood} employs node dropping and attribute masking for contrastive bootstrapping~\cite{grill2020bootstrap}. MoleOOD~\cite{yang2022learning} uses VI to guide invariant learning (IL). LiSA~\cite{yu2023mind} generates variational subgraphs under information constraints and encourages diversity via energy-based regularization. EERM~\cite{wu2022handling} simulates adversarial virtual environments for node-level IL.

\subsubsection{Adversarial Learning Approaches}

Adversarial learning (AL) has been widely utilized for OOD generalization~\cite{yi2021improved} to reduce the domain discrepancy, which is naturally extended to graph data. 
Some approaches utilize AT to generate effective perturbation for enhancing the generalization capacity. 
For example, GraphAT~\cite{feng2019graph} adds learnable perturbations to the target graph, which are trained to degrade the smoothness, addressing the worst-case problem. Yet it underutilizes the graph structure and environmental information. 
CAP~\cite{xue2021cap} maximizes the training loss in the neighborhood of model parameters and node attributes, which can mitigate the risk of falling into the local minima. 
AIA~\cite{sui2023unleashing} generates augmented data by merging dual masks for environmental/stable features, preserving semantics. Its regularization terms constrain perturbations to ensure optimization stability. 
LECI~\cite{gui2023joint} adopts causal analysis from~\cite{chen2022learning} to remove spurious correlations. It uses AT to enforce subgraph independence from labels/environments via an discriminator. 
WT-AWP~\cite{wu2023adversarial} adapts the adversarial weight perturbation to graph classification as a regularization term, which is applied on partial layers to relieve potential gradient vanishing. 
Several domain adaption approaches also utilize AT to align graph representations across different domains~\cite{yin2022deal,wu2020unsupervised}. 
DEAL~\cite{yin2022deal} leverages adversarial perturbation on node attributes to transfer source graphs into the target domain. 
In summary, these AT approaches can implicitly reduce the distribution discrepancy in the embedding space across domains. However, they usually require prior knowledge of domain or environment labels.

\subsubsection{Discussion}
% Beyond the development of scoring functions for OOD detection on graphs, recent studies have further extended this line of research towards enhanced explainability and generalization. In particular, for graph-level OOD detection and anomaly detection, several methods provide meaningful interpretations for the detection results. For instance, SIGNET~\cite{liu2023towards} proposes a self-interpretable framework that simultaneously outputs anomaly scores and explanatory subgraphs by maximizing mutual information across multi-view subgraphs. SGOOD~\cite{ding2023sgood} incorporates substructure information and their relationships to learn substructure-enhanced graph representations for OOD detection.

% In parallel, OOD generalization techniques have also been integrated into graph learning. Besides the above works, more techniques on OOD generalization have also been extended to this problem including CL~\cite{liu2023flood,chen2022learning,luo2023rignn,luo2023towards_graph} and representation decorrelation (RD)~\cite{li2022ood,qin2022graph}. 
% The performance of OOD link prediction is also studied through theoretical analyses~\cite{zhou2022ood}. 
% To facilitate the research on this crucial problem, two extensive benchmarks have been constructed~\cite{ji2022drugood,gui2022good}. These problems have also been applied to more practical scientific research including molecular property prediction~\cite{yang2022learning} as well as drug discovery~\cite{ji2023drugood}.

Beyond designing scoring functions for OOD detection on graphs, recent efforts have emphasized explainability and generalization. For graph-level OOD and anomaly detection, methods such as SIGNET~\cite{liu2023towards} jointly produce anomaly scores and explanatory subgraphs by maximizing mutual information across multi-view subgraphs, while SGOOD~\cite{ding2023sgood} leverages substructure information for enhanced representations. 
Meanwhile, generalization techniques like CL~\cite{liu2023flood,chen2022learning,luo2023rignn,luo2023towards_graph} and representation decorrelation (RD)~\cite{li2022ood,qin2022graph} have been adapted to improve OOD robustness. Theoretical studies also explore OOD link prediction~\cite{zhou2022ood}, and large-scale benchmarks~\cite{gui2022good} have been developed to support empirical evaluation. These advances have been applied in real-world domains such as molecular property prediction and drug discovery~\cite{yang2022learning}.

%% file: 9_future_work.tex
\section{Conclusion and Future Work}

% In summary, this paper provides a comprehensive overview of how real-world GNNs address four key challenges: \emph{imbalance, noise, privacy, and OOD}. These aspects are often overlooked in most literature reviews. We begin by discussing the vulnerabilities and limitations of existing GNN models, revealing crucial challenges. Subsequently, we meticulously introduce the frameworks and principles of existing GNN models for addressing each key factor, categorizing them in detail. We also highlight the key technological contributions of representative works and conclude with some exploratory discussions. Despite significant progress in addressing real-world GNNs, there remain promising directions for future research in this field, which we further analyze here:
In summary, this survey presents a comprehensive review of how real-world GNNs tackle four major challenges: \emph{imbalance, noise, privacy, and OOD}, which are often underrepresented in prior surveys. We first discuss the vulnerabilities and limitations of current models to reveal critical challenges, followed by a detailed categorization of existing methods addressing each aspect. Representative works are highlighted for their key contributions. In the following, we conclude with forward-looking discussions on promising future directions in real-world GNN research.

% \smallskip\textbf{Enhancing Scalability.}
% Existing researches mainly focus on the imbalance, noise, privacy, and OOD issues in small-scale graph datasets, leaving a considerable gap with the more prevalent large-scale graph datasets in the real world. These problems become more intricate within massive large-scale graphs, demanding higher performance and efficiency from model design. For instance, G$^2$GNN~\cite{wang2022imbalanced} mitigates the imbalance issue by calculating graph similarity using graph kernels to construct a graph of graphs, but the usage of graph kernel and pairwise similarity computation limits its applicability to massive and large-scale graphs. Additionally, exploring pre-training on small-scale graphs and then generalizing to large-scale graphs that are imbalanced, noisy, or out of distribution is an interesting and significant direction.
\smallskip\textbf{Enhancing Scalability.}
Most existing studies address imbalance, noise, privacy, and OOD challenges on small-scale graph datasets, leaving a gap with the large-scale graphs common in real-world scenarios. These issues become more complex at scale, requiring models with greater efficiency and robustness. For example, G$^2$GNN~\cite{wang2022imbalanced} alleviates imbalance by constructing a graph-of-graphs using graph kernel-based similarity, but its reliance on pairwise computations limits scalability. Exploring pre-training on small graphs and transferring to large, imbalanced, noisy, or OOD graphs remains a promising and impactful direction.

% \smallskip\textbf{Improving Interpretability.}
% Many real-world applications of GNNs, such as drug discovery, medical decision making, and traffic planning, demand high interpretability from models. Despite a range of existing approaches that have achieved great performance in some real-world scenarios like class imbalance and OOD generalization, the exploration of models' interpretability remains limited. Providing explanations or decision processes along with prediction results is crucial for enhancing the reliability of models and defending from attacks~\cite{zhang2021backdoor}. For instance, SIGNET~\cite{liu2023towards} infers both graph-level anomaly scores and subgraph-level explanations by maximizing the mutual information of constructed multi-view subgraphs, providing a reliable graph-level anomaly detection framework. Integrating techniques like built-in interpretability~\cite{zhang2022protgnn}, post-hoc explanations~\cite{ying2019gnnexplainer}, causal discovery~\cite{wu2021discovering} and counterfactual explanations~\cite{bajaj2021robust} into real-world GNN models is a promising research direction that can serve as an assurance for their applications in critical and private scenarios.
\smallskip\textbf{Improving Interpretability.}
Many real-world GNN applications, such as drug discovery, healthcare, and traffic planning, require high interpretability. Although existing methods have demonstrated strong performance under challenges like class imbalance and OOD generalization, interpretability remains underexplored. Enhancing model transparency through explanations is vital for reliability and robustness against attacks~\cite{zhang2021backdoor}. For example, SIGNET~\cite{liu2023towards} jointly outputs anomaly scores and explanatory subgraphs via multi-view mutual information maximization. Incorporating built-in interpretability, post-hoc explanation, and counterfactual reasoning offers a promising path toward trustworthy GNNs in sensitive domains.

% \smallskip\textbf{More Theoretical Guarantees.}
% Establishing theoretical guarantees is crucial for developing reliable real-world GNN models. However, the previous theoretical understanding of GNNs primarily focuses on their expressive power~\cite{xu2018powerful,chen2019equivalence}, while theoretical guarantees for GNNs' generalization in complex real-world scenarios such as noise disturbance and OOD generalization are still underexplored. Theoretical analysis for these scenarios can validate GNN models' ability to handle natural interferences or deliberate attacks, facilitating their deployment in safety-critical applications. For instance, recently, GraphGuard~\cite{anonymous2024graphguard} provides theoretical defense guarantees against perturbations in graph structure and node features for graph classification, demonstrating the model's reliability against a limited number of attacks. Investigating theoretical guarantees for more real-world scenarios, such as class imbalance and label noise, and further providing a unified theoretical analysis framework, is significant for the broader application of GNNs in critical real-world contexts.
\smallskip\textbf{More Theoretical Guarantees.}
Establishing theoretical guarantees is essential for building reliable GNNs in real-world settings. While prior work has mainly focused on expressive power~\cite{chen2019equivalence}, theoretical insights into GNNs' generalization under noise, distribution shifts, and adversarial conditions remain limited. Such analyses can validate GNNs' robustness to natural perturbations and adversarial attacks, supporting their deployment in safety-critical applications. For example, GraphGuard~\cite{anonymous2024graphguard} offers provable robustness against structural and feature perturbations in graph classification. Advancing theoretical guarantees is vital under scenarios like class imbalance and label noise, and developing unified analytical frameworks.

% \smallskip\textbf{Comprehensive Benchmarks and Universal Models.}
% These real-world scenarios are often individually studied, and seldom considered comprehensively. Existing models are mostly designed and optimized for specific scenarios, performing well in those contexts but failing in many other scenarios. 
% For instance, UDA-GCN~\cite{wu2020unsupervised}, which is designed for graph domain adaptation, fails in graph transfer learning with noisy labels, a more complex and realistic scenario~\cite{yuan2023alex}. 
% Therefore, proposing a comprehensive real-world graph benchmark is necessary, which can systematically evaluate a model's ability to address various real-world challenges and provide an integrated score. This benchmark can significantly advance the development and fair comparison of graph neural network models in real-world applications. Simultaneously, leveraging this benchmark, to develop a universally applicable and robust model with a high integrated score is a very promising direction.
\smallskip\textbf{Comprehensive Benchmarks and Universal Models.}
Real-world scenarios are often studied in isolation, with existing models tailored to specific settings, limiting their generalizability. For example, UDA-GCN~\cite{wu2020unsupervised}, designed for domain adaptation, fails under more complex tasks such as transfer learning with noisy labels. So, a comprehensive benchmark is needed to systematically assess models across diverse real-world challenges and yield an integrated score. Such a benchmark would facilitate fair evaluation and drive progress toward developing universally robust GNNs capable of performing well across heterogeneous conditions.

% \smallskip\textbf{Towards More Realistic Applications.}
% Developing more realistic GNN models is important for more real-world applications in diverse fields. For example, GNNs are used to analyze protein-protein interaction networks~\cite{reau2023deeprank}. These networks can be highly imbalanced, with some proteins being more common than others. GNNs help in identifying novel interactions, which is crucial for understanding diseases and developing new drugs.
% Moreover, when applied to new organisms or under-studied proteins, the models might face OOD data. 
% In the financial domain, GNNs are effective in detecting fraudulent transactions in large and complex financial networks~\cite{liu2021pick}. Fraudulent transactions are typically rare (class imbalance) and can appear in OOD patterns.
% GNNs can analyze road networks to optimize routes for autonomous vehicles~\cite{yu2020learning}. The variability in traffic conditions and road closures presents OOD challenges.
% Therefore, realistic GNN models are highly expected to adapt to new traffic data and changing environments by employing online learning techniques.
\smallskip\textbf{Towards More Realistic Applications.}
Building realistic GNN models is critical for broader deployment in domains such as biology, finance, and transportation. For instance, GNNs aid in analyzing protein-protein interaction networks~\cite{reau2023deeprank}, which are often imbalanced and exhibit OOD behavior when applied to new organisms. In finance, GNNs effectively detect rare fraudulent transactions in large-scale networks~\cite{liu2021pick}, while in transportation, they support route optimization under dynamic conditions~\cite{yu2020learning}. These applications highlight the need for GNNs that can handle class imbalance, adapt to OOD data, and incorporate online learning to manage evolving environments.